\documentclass{article}
\usepackage{amsmath,inputenc}

\title{Online Learning for Traffic Routing under Unknown Preferences}
\author{Devansh Jalota$^\dagger$, Karthik Gopalakrishnan$^\dagger$, Navid Azizan$^\ddagger$, \\ Ramesh Johari$^\dagger$, and Marco Pavone$^\dagger$%
\thanks{$^\dagger$ Stanford University; {\tt \{djalota, kgopalakrishnan, rjohari, pavone\}@stanford.edu}.}%
\thanks{$^\ddagger$ Massachusetts Institute of Technology; {\tt azizan@mit.edu}.}%
}

\newif\ifarxiv   
\arxivtrue 

\ifarxiv
\else
\fi

\date{March 2022}

\usepackage{graphicx}
\usepackage{csvsimple}
\usepackage{rotating}
\usepackage[letterpaper, portrait, margin=1in]{geometry}
\usepackage{hyperref}
\hypersetup{
    colorlinks=true,
    linkcolor=blue,
    citecolor=red,
    filecolor=magenta,      
    urlcolor=cyan
    }

\usepackage[nocomma]{optidef}
\usepackage{listings}
\usepackage{comment}
\usepackage{appendix}
\usepackage[ruled,vlined]{algorithm2e}
\usepackage{amsfonts} 
\usepackage{amssymb}
\usepackage{bm}
\usepackage[square,comma,numbers]{natbib}

\usepackage{tikz}
\usepackage{pgfplots}
\DeclareUnicodeCharacter{2212}{−}
\usepgfplotslibrary{groupplots,dateplot}
\usetikzlibrary{patterns,shapes.arrows}
\pgfplotsset{compat=newest}

\usepackage{caption}
\usepackage{subcaption}

\usepackage{color} 
\definecolor{mygreen}{RGB}{28,172,0} 
\definecolor{mylilas}{RGB}{170,55,241}
\DeclareFixedFont{\ttb}{T1}{txtt}{bx}{n}{12} 
\DeclareFixedFont{\ttm}{T1}{txtt}{m}{n}{12}  
\usepackage{bbm}
\usepackage{amsmath,amsthm}

\newtheorem{theorem}{Theorem}

\newtheorem{lemma}{Lemma}

\theoremstyle{definition}
\newtheorem{definition}{Definition}

\newenvironment{hproof}{%
  \proof}{\endproof}

\newcommand{\norm}[1]{\left\lVert#1\right\rVert}

\usepackage{color}
\definecolor{deepblue}{rgb}{0,0,0.5}
\definecolor{deepred}{rgb}{0.6,0,0}
\definecolor{deepgreen}{rgb}{0,0.5,0}

\usepackage{listings}

\renewcommand\footnotemark{}
\def\x{\bm{x}}
\def\f{\bm{f}}

\def\v{\bm{v}}
\def\p{\bm{p}}

\def\c{\bm{c}}

\def\l{\bm{l}}
\def\w{\bm{w}}
\def\ppi{\bm{\pi}}
\def\P{\mathcal{P}}

\def\D{\mathcal{D}}
\def\U{\mathcal{U}}
\def\0{\bm{0}}

\def\ttau{\bm{\tau}}

\def\ttau{\boldsymbol{\tau}}

\DeclareMathOperator*{\argmin}{arg\,min}

\begin{document}

\maketitle

\begin{abstract}
\ifarxiv
In transportation networks, users typically choose routes in a decentralized and self-interested manner to minimize their individual travel costs, which, in practice, often results in inefficient overall outcomes for society. As a result, there has been a growing interest in designing road tolling schemes to cope with these efficiency losses and steer users toward a system-efficient traffic pattern. However, the efficacy of road tolling schemes often relies on having access to complete information on users' trip attributes, such as their origin-destination (O-D) travel information and their values of time, which may not be available in practice.

\fi

Motivated by this practical consideration, we propose an online learning approach to set tolls in a traffic network to drive heterogeneous users with different values of time toward a system-efficient traffic pattern. In particular, we develop a simple yet effective algorithm that adjusts tolls at each time period solely based on the observed aggregate flows on the roads of the network without relying on any additional trip attributes of users, thereby preserving user privacy. In the setting where the O-D pairs and values of time of users are drawn i.i.d. at each period, we show that our approach obtains an expected regret and road capacity violation of $O(\sqrt{T})$, where $T$ is the number of periods over which tolls are updated. Our regret guarantee is relative to an offline oracle that has complete information on users' trip attributes. We further establish a $\Omega(\sqrt{T})$ lower bound on the regret of any algorithm, which establishes that our algorithm is optimal up to constants. Finally, we demonstrate the superior performance of our approach relative to several benchmarks on a real-world transportation network, thereby highlighting its practical applicability.
\end{abstract}

\section{Introduction}

Many real-world systems are composed of self-interested users that interact non-cooperatively in a shared environment. However, in such systems, the lack of coordination between self-interested users who seek to optimize their objectives often results in inefficient outcomes that are at odds with the goals of the system designer. For instance, in transportation networks, the \emph{selfish routing} by users who choose routes to minimize their travel times~\cite{ROUGHGARDEN2004389} typically results in a traffic pattern that is far from an efficient traffic assignment~\cite{Sheffi1985}. As a result, there has been a growing interest in designing intervention and control schemes to cope with the selfishness of users~\cite{how-bad-is-selfish} across a wide range of resource allocation applications, including energy management in smart grids\ifarxiv~\cite{azizan2020optimal,Palensky11demandside,azizan2018opportunities} \fi and traffic routing on road networks~\cite{pigou,Sharon2018TrafficOF}. One promising method that has emerged to mitigate efficiency losses is to set prices on the shared resources to influence and steer user behavior to align with system-efficient outcomes\ifarxiv~\cite{cole2003pricing,ZPCP-2018}\fi.

In the context of traffic routing, road tolls are a commonly used mechanism to cope with the inefficiency loss due to the selfishness of users and enforce the system-optimum solution as a user equilibrium~\cite{karakostas2004edge}. However, the computation of these tolls typically relies on solving a centralized optimization problem, which assumes complete information on users' trip attributes\ifarxiv~\cite{fleisher-tolls,jalota2022balancing,jalota-cprr}\fi, such as their origin-destination (O-D) travel information and their values of time~\cite{walters1961theory}. In practice, this information is typically not available since this would violate user privacy and can thus confound the successful deployment of a tolling scheme to regulate road traffic. Furthermore, the trip attributes of users often vary with time, e.g., when users' values of time and O-D pairs are drawn i.i.d. from some unknown distribution. As a result, the central planner may need to periodically collect users' trip attributes and re-solve a large scale optimization problem that may be computationally expensive.



As a result, in this work, we propose an online learning approach to set tolls in a traffic network to steer heterogeneous users with different values of time to a system optimum traffic pattern that minimizes the sum of the travel times of all users weighted by their values of time. In this setting, we assume that the value of time and O-D pair of each user are private information and cannot be used to design optimal tolling policies. 
Our method to set road tolls is different from prior centralized optimization approaches that require complete information on the values of time and O-D pairs of users. In this incomplete information setting, our algorithmic approach \ifarxiv to setting tolls \fi relies on adjusting road tolls based on the observed aggregate flows on the edges of the \ifarxiv traffic \fi network. We mention that such aggregate flow data is readily available through modern sensing technologies, such as loop detectors, and helps maintain user privacy.

\ifarxiv
\subsection{Our Contributions}
\fi
In this work, we study the problem of setting optimal tolls that minimize the total system cost, i.e., the sum of the travel times of all users weighted by their values of time, in a capacity-constrained traffic network. We study this problem in the incomplete information setting when the O-D pairs and values of time of heterogeneous users are not known. Since centralized optimization approaches are typically not conducive in this setting, we consider the problem of learning tolls over time to minimize (i) regret, i.e., the optimality gap between the resulting allocation and that of an offline oracle with complete information on users' trip attributes, 
and (ii) constraint violation, i.e., the extent to which the road capacity constraints are violated.


To this end, we develop a simple yet effective approach to set tolls that preserves user privacy while achieving sub-linear regret and constraint violation guarantees in the number of periods $T$ over which the tolls are updated. \ifarxiv In particular, our algorithmic approach adjusts tolls solely based on the observed aggregate flows on the roads of the traffic network and is akin to performing gradient descent on the dual of a system optimization problem (see Section~\ref{sec:sysopt}). \else \fi Furthermore, we establish a regret lower bound to show that our algorithm achieves the optimal regret guarantee, up to constants.


We then evaluate the performance of our approach on a real-world transportation network. The results of our experiments not only validate the theoretical regret and constraint violation guarantees but also highlight the superior performance of our algorithm relative to several natural benchmarks. 
\ifarxiv We further demonstrate the practical efficacy of our approach by showing that it achieves a total travel time close to the minimum achievable total travel time in the network. Finally, we investigate several properties of the computed tolls to demonstrate that our proposed tolling schemes align with typical congestion pricing policies in real-world transportation networks. \fi

\ifarxiv
\paragraph{Organization:} 
Our paper is organized as follows. Section~\ref{sec:litreview} reviews related literature. We then present a model of traffic flow as well as the regret and constraint violation performance measures to evaluate the efficacy of a tolling policy in Section~\ref{sec:model}. Then, in Section~\ref{sec:mainResult}, we introduce our online learning algorithm and its associated regret and constraint violation guarantees. Next, we evaluate the performance of our approach on a real-world transportation network through numerical experiments in Section~\ref{sec:expt}. Finally, we conclude the paper and provide directions for future work in Section~\ref{sec:conclusion}.
\fi

\section{Literature Review} \label{sec:litreview}

\ifarxiv
Resource allocation under heterogeneous user preferences has been a widely studied topic in the computer science, operations research, and economics communities with a wide range of applications, including traffic routing, smart-grid management, and public-good allocation. Traditional approaches to achieve an efficient allocation of resources have typically relied on having access to complete information on the preferences of the different users. For instance, in traffic routing applications, where users have differing values-of-time, centralized resource allocation methods have been used to design congestion pricing schemes that induce system efficient traffic routing patterns as equilibria~\cite{dafermos-multiclass,HAN2008753,karakostas2004edge}. In the context of power flow problems,~\cite{TG-2014,ne-Design-Coord-2009} assume knowledge of the reward functions of the different players to distributedly converge to the desired Nash equilibria. Furthermore, in the context of allocating capacity-constrained public goods to users,~\cite{Jalota2020MarketsFE,jalota2021fisher} develop methods based on Fisher markets to allocate resources given complete access to user's preferences. However, these approaches are, in general, practically infeasible since having complete information on users' preferences would violate user privacy.
\fi


As a result, there has been a growing interest in designing mechanisms that do not require complete information of users' preferences to achieve an efficient resource allocation. To this end, mechanism design has enabled the truthful elicitation of users' private information\ifarxiv~\cite{NIPS2004_fc03d482,heydaribeni2018distributed,pmlr-v119-deng20d}\fi. Furthermore, inverse game theory~\cite{igt-main} has enabled the learning of users' preferences, e.g., O-D travel demands\ifarxiv~\cite{WSZP-2019,data-inverse-opt}\fi in traffic routing. In contrast to these works, we do not directly learn or elicit user preferences to set optimal tolls. Instead, our approach bypasses the need to have information on users' preferences by using the total observed flow on each road as feedback to update tolls while retaining good performance.

\ifarxiv
Our toll update procedure is, in principle, similar to price update mechanisms that utilize past observations of user consumption behavior, rather than the knowledge of users' preferences, to inform future pricing decisions. For instance, Kleinberg and Leighton~\cite{kleinberg2003value} consider a revenue maximization problem setting and use information from interactions with earlier buyers to inform pricing decisions for subsequent buyers under several informational assumptions on the valuations of users. In the context of online linear programming, Li et al.~\cite{li2020simple} develop an algorithm for binary linear programs that updates prices of goods at each time step based on the observed demand of a user at the previous time step. The algorithm in Li et al.~\cite{li2020simple} is analogous to Walrasian tatonnement-based approaches, which adjust prices based on the discrepancy between the supply and demand of every good~\cite{ausubel-dynamic-auction}. Along a similar vein, Bistritz and Bambos~\cite{bistritz2021online} develop an online learning approach in load balancing problems wherein the prices are updated based solely on the total load on each resource and show that this algorithm converges to a load-balanced Nash equilibrium. 
\fi

Dynamic pricing mechanisms have also been studied in the incomplete information setting for traffic routing applications. For instance,\ifarxiv~\cite{YANG2004477,YANG2010157,WANG20121085}\fi study trial-and-error approaches to set marginal cost tolls in the absence of users' demand functions. Furthermore,~\cite{melo-2011} develops dynamic variants of Pigou's solution to mitigate negative externalities by nudging users toward the system-optimum solution. In line with the above approaches, we also learn and iteratively update road tolls through repeated interactions of users in the traffic network. However, in contrast to these approaches that typically update marginal cost tolls in the setting of homogeneous users, we use linear programming duality to update tolls for heterogeneous users with different values of time\ifarxiv~\cite{HAN2008753,dafermos-multiclass,karakostas2004edge} \else \fi.




\section{Model and Problem Formulation} \label{sec:model}

In this section, we introduce the basic definitions and user behavior model we consider in this work (Section~\ref{sec:def-usermodel}), the system optimization objective of the central planner 
(Section~\ref{sec:sysopt}), and the performance measures to evaluate the efficacy of a tolling policy (Section~\ref{sec:perf-measures}).

\subsection{Preliminaries and User Optimization} \label{sec:def-usermodel}

We study the problem of routing heterogeneous users between their respective origins and destinations in a capacity-constrained road network. The road network is modeled as a directed graph $G = (V, E)$, with vertex and edge sets denoted by $V$ and $E$, respectively. Each edge $e \in E$ has a fixed capacity $c_e$ and a fixed latency (travel time) $l_e$, and we denote $\c = \{c_e \}_{e \in E}$ as the vector of edge capacities\ifarxiv and $\l = \{l_e \}_{e \in E}$ as the vector of edge travel times.\fi 

The set of all users is denoted by $\U$ and each user $u \in \U$ makes trips in the road network between an origin-destination (O-D) pair $w_u = (s_u, t_u)$, where $s_u$ represents the origin and $t_u$ represents the destination of the trip. Each O-D pair is connected by paths, which are finite sequences of directed edges \ifarxiv beginning at the origin and ending at the destination, \fi and each user $u$ selects one such path from the set of all paths $\P_{u}$ that connect the O-D pair $w_u$. 
An assignment of users to paths is denoted by the vector $\f = \{f_{P, u}: P \in \P_{u}, u \in \U \}$, where $f_{P,u}\in \{ 0, 1\}$ denotes whether user $u$ is assigned to path $P$. Each path flow $\f$ corresponds to a unique edge flow vector $\x = \{x_e:e\in E \}$, where \ifarxiv the edge flow \fi $x_e = \sum_{u \in \U} \sum_{P \in \P_u: e \in P} f_{P,u}$. Each user $u$ also has an \emph{outside option} \ifarxiv of not using the network \fi for which the user incurs a cost $\lambda_u$, which can be interpreted as the cost for not completing the trip, e.g., staying at home. The outside option is a commonly used modelling assumption in the transportation literature~\cite{nikzad2017thickness,ostrovsky2019carpooling}, and, in this traffic routing context, captures a decision making framework where each user $u$ may not wish to incur a cost, including travel time and tolls, of more than $\lambda_u$ for their trip. 
We denote whether users choose the outside option through the vector $\f_o = \{ f_{o,u}: u \in \U \}$, where $f_{o,u}\in \{ 0, 1\}$ is a binary variable that takes the value one when user $u$ chooses the outside option.


Users are assumed to be selfish and thus choose a path (or the outside option) that minimizes their travel cost, which we assume is a linear function of tolls and travel time. For each user \ifarxiv $u \in \U$ \fi with a value-of-time $v_u>0$ and a vector of edge prices (or tolls) $\boldsymbol{\tau} = \{\tau_e\}_{e \in E}$, the travel cost on a path $P$ is given by $C_{P}(\boldsymbol{\tau}) = \sum_{e \in P} \left( v_u l_e +  \tau_e \right)$. For ease of notation, we denote the total travel time on path $P$ as $l_P = \sum_{e \in P} l_e$ and its toll as $\tau_P = \sum_{e \in P} \tau_e$. 
Then, given a vector of tolls $\ttau$, we define a path flow $\f$ and outside option flow $\f_o$ to be an \emph{equilibrium} if each user chooses a path (or the outside option) that minimizes their travel cost.

\begin{definition} [Equilibrium] \label{def:eqnotion}
For a given vector of tolls $\ttau$, a path flow $\f$ and outside option flow $\f_o$ is an equilibrium if for each user $u$, $\sum_{P \in \mathcal{P}_{u}} f_{P,u} + f_{o,u} = 1$, where
$f_{P_u^*,u} = 1$ for some path $P_u^*$ if
\ifarxiv
\begin{align*}
    C_{P_u^*}(\ttau) \leq \min \big\{ \min_{Q\in \P_u} \{ C_Q(\ttau) \}, \lambda_u \big\}
\end{align*}
or $f_{o,u} = 1$ for the outside option if
\begin{align*}
    \lambda_u \leq C_{P}(\ttau), \text{ for all } P \in \P_u.
\end{align*}
\fi
\end{definition}
\ifarxiv
A few comments about our modeling assumptions and Definition~\ref{def:eqnotion} are in order. First, under this equilibrium notion, all users route their entire flow on one path (or the outside option), which is analogous to atomic equilibria~\cite{roughgarden_2007} in congestion games. Next, note that under our modeling assumptions, the travel time on any edge is independent of the number of users on that edge as long as the total number of users on that edge does not exceed its capacity. This approximation is largely consistent with observed travel times on real-world road networks, where the travel times on roads tend to stay relatively constant up until the road capacity, beyond which the travel time increases steeply~\cite{li2011fundamental}. For a more detailed discussion on the validity of this assumption in modeling real-world traffic networks, we refer to\ifarxiv~\cite{varaiya2005we,ostrovsky2019carpooling}\else~\cite{ostrovsky2019carpooling}\fi. Finally, observe that under an equilibrium flow all users take the shortest ``cost'' path (or the outside option) with respect to the set tolls $\ttau$ irrespective of whether this violates the road capacity constraints. In practice, setting the tolls too low will likely result in road capacity violations and thus congestion delays. However, as with general equilibrium models that do not model users' responses to shortages in the supplies of goods, for our purposes, we do not model traffic congestion since we seek to set tolls that keep capacity violations small. We note that in road networks small levels of capacity violation are acceptable, since road capacities serve as a ball-park for the number of vehicles on the road such that congestion delays do not increase significantly.
\fi

\subsection{System Optimization and Efficient Tolls} \label{sec:sysopt}

In this section, we present the system optimization problem of the central planner and the notion of market-clearing tolls that induce the optimal solution to the system optimization problem as an equilibrium flow.

\ifarxiv
\paragraph{System Optimization:} 
We now present the problem faced by the central planner, who seeks to minimize the total system cost 
while ensuring that the resulting traffic assignment is feasible.
\fi
\begin{mini!}|s|[2]                   
    {\f, \f_o}                              
    {U^* = \sum_{u \in \U} \Big( v_u \sum_{P \in \mathcal{P}_u }  l_P f_{P,u} + \lambda_u f_{o,u} \Big), {\label{eq:capacityObj1IP}} } 
    {\label{eq:Eg001}}             
    {}                                
    \addConstraint{\sum_{P \in \mathcal{P}_{u}} f_{P,u} + f_{o,u}}{= 1, \quad \forall u \in \U, } {\label{eq:capacityCon1IP}}   
    \addConstraint{\f_o}{\in \{0, 1 \}^{|U|}, f_{P,u} \in \{0,1 \}, \ifarxiv \quad \fi \forall P \in \P_u, u \in \U, } {\label{eq:capacityCon2IP}}
    \addConstraint{\sum_{u \in \U} \sum_{P \in \mathcal{P}_{u} : e \in P} f_{P,u}}{  \leq c_e, \quad \forall e \in E , } {\label{eq:capacityCon3IP}}
\end{mini!}
Here,~\eqref{eq:capacityObj1IP} is the total system cost objective, i.e., the sum of the travel times of all users weighted by their values of time for all users using the network and the cost of the outside option for all users who do not use the network. Furthermore,~\eqref{eq:capacityCon1IP} are user allocation constraints since users will either use one of a set of feasible paths or the outside option,~\eqref{eq:capacityCon2IP} are binary allocation constraints, and~\eqref{eq:capacityCon3IP} are the edge capacity constraints. Note that a feasible solution to Problem~\eqref{eq:capacityObj1IP}-\eqref{eq:capacityCon3IP} exists since it is feasible for all users to choose the outside option $o$.

\ifarxiv
\paragraph{Efficient Tolls:} 
\fi

The central planner is tasked with setting tolls $\ttau$ that induce the system optimum solution, i.e., the solution to Problem~\eqref{eq:capacityObj1IP}-\eqref{eq:capacityCon3IP}, as an equilibrium flow. 
One such set of tolls that achieve this goal are \emph{market-clearing} tolls that induce equilibrium flows $\f, \f_o$ satisfying the road capacity constraints, i.e., $\sum_{u \in \U} \sum_{P \in \P_u: e \in P} f_{P,u} = x_e \leq c_e$ for all edges $e \in E$. Additionally, market-clearing tolls satisfy the property that $\tau_e = 0$ for all edges $e$ such that the edge flow $x_e < c_e$, and $\tau_e \geq 0$ for all edges $e$ such that $x_e = c_e$. Theorem~\ref{thm:eff-result} establishes that if the tolls are market clearing then the resulting equilibrium flows minimize the total system cost, i.e., the corresponding equilibrium flow is an optimal solution to Problem~\eqref{eq:capacityObj1IP}-\eqref{eq:capacityCon3IP}.

\ifarxiv
\begin{theorem} [Efficiency of Market-Clearing Tolls] \label{thm:eff-result}
Suppose that the tolls $\ttau^*$ are market-clearing. Then, the corresponding equilibrium flows, given by a path flow $\f^*$ and outside option flow $\f_o^*$, are optimal solutions to Problem~\eqref{eq:capacityObj1IP}-\eqref{eq:capacityCon3IP}.
\end{theorem}
\fi

\ifarxiv
\begin{proof}
To prove this claim, we show that the equilibrium flows $\f^*, \f_o^*$ corresponding to the toll $\ttau^*$ minimizes the total system cost, i.e., $\sum_{u \in \U} \left( v_u \sum_{P \in \mathcal{P}_u }  l_P f_{P,u}^* + \lambda_u f_{o,u}^* \right) \leq \sum_{u \in \U} \left( v_u \sum_{P \in \mathcal{P}_u }  l_P f_{P,u} + \lambda_u f_{o,u} \right)$ for any other feasible flows $\f, \f_o$ that satisfy Constraints~\eqref{eq:capacityCon1IP}-\eqref{eq:capacityCon3IP}. To this end, suppose, without loss of generality, that under the equilibrium flows corresponding to the toll $\ttau^*$, users $u \in \U_1$ choose paths $P_1^*, \ldots, P_k^*$, i.e., $f_{P_u^*, u}^* = 1$ for some path $P_u^* \in \P_u$ for all users $u \in \U_1$, while users $u \in \U_2$ choose the outside option, i.e., $f_{o,u}^* = 1$ for all users $u \in \U_2$, where $\U_1 \cup \U_2 = \U$ and $\U_1 \cap \U_2 = \emptyset$. Then, for any other feasible flows $\f, \f_o$ that satisfy the Constraints~\eqref{eq:capacityCon1IP}-\eqref{eq:capacityCon3IP}, it holds by the definition of an equilibrium for all users $u \in \U_1$ that
\begin{align*}
    &v_u l_{P_u^*} + \sum_{e \in P_u^*} \tau_e^* \leq \sum_{P \in \mathcal{P}_u } \left( v_u  l_P + \sum_{e \in P} \tau_e^* \right) f_{P,u} + \lambda_u f_{o,u},
\end{align*}
and for users $u \in \U_2$ that
\begin{align*}
    \lambda_u \leq \sum_{P \in \mathcal{P}_u } \left( v_u  l_P + \sum_{e \in P} \tau_e^* \right) f_{P,u} + \lambda_u f_{o,u}.
\end{align*}
Summing up the above inequalities for all users $u \in \U$ and rearranging the equation we get that
\begin{align*}
    \sum_{u \in \U_1} v_u l_{P_u^*} + \sum_{u \in \U_2} \lambda_u &\leq \sum_{P \in \mathcal{P}_u } \left( v_u  l_P + \sum_{e \in P} \tau_e^* \right) f_{P,u} + \lambda_u f_{o,u} - \sum_{u \in \U_1} \sum_{e \in P_u^*} \tau_e^*.
\end{align*}
Finally, since the tolls $\ttau^*$ are market-clearing, it holds that $\sum_{u \in \U_1} \sum_{e \in P_u^*} \tau_e^* = \sum_{e \in E} \tau_e^* c_e \geq \sum_{e \in E} \tau_e^* x_e$ for any feasible edge flow $\x \leq \c$. Thus, it follows that
\begin{align*}
    \sum_{u \in \U} \left( v_u \sum_{P \in \mathcal{P}_u }  l_P f_{P,u}^* + \lambda_u f_{o,u}^* \right) &= \sum_{u \in \U_1} v_u l_{P_u^*} + \sum_{u \in \U_2} \lambda_u, \\ 
    &\leq \sum_{u \in \U} \left( \sum_{P \in \mathcal{P}_u } \left( v_u  l_P + \sum_{e \in P} \tau_e^* \right) f_{P,u} + \lambda_u f_{o,u} \right) - \sum_{e \in E} \tau_e^* c_e, \\
    &\leq  \sum_{u \in \U} \left( v_u \sum_{P \in \mathcal{P}_u } l_P f_{P,u} + \lambda_u f_{o,u} \right),
\end{align*}
which proves our claim.
\end{proof}
\fi \ifarxiv Theorem~\ref{thm:eff-result} establishes that if the central planner set market-clearing tolls, then the corresponding equilibrium flows would not only satisfy the Constraints~\eqref{eq:capacityCon1IP}-\eqref{eq:capacityCon3IP} but also result in the lowest total system cost among all feasible assignments. \fi While market-clearing tolls provide a method to induce equilibrium flows that minimize the total system cost, such tolls cannot typically be computed, e.g., through linear programming (see Section~\ref{sec:lpdual}), since the values of time and O-D pairs of users are, in general, unknown to the central planner. Furthermore, these user specific attributes tend to be time-varying, e.g., when users' values of time and O-D pairs are drawn i.i.d. from some distribution. Thus, a central planner would need to periodically collect these parameters and re-solve a large optimization problem to update the tolls\ifarxiv at each step\fi, which may not be practically viable.

As a result, in this work, we consider an online learning approach that only relies on past observations of aggregate flows on each road of the network to set tolls and steer heterogeneous users toward a system-optimum traffic pattern over time. Our motivation for pursuing an online learning method is three-fold. First, as opposed to centralized approaches, an online learning approach can bypass the need to have complete information on users' preferences. Next, mechanism design approaches that rely on truthfully eliciting users' preferences may not be practically viable and often be insufficient in inducing the system-optimum solution as an equilibrium flow \ifarxiv(see Appendix~\ref{apdx:vcg})\fi. Finally, modern sensing technologies, e.g., loop detectors, are well equipped to collect aggregate road flow data and thus do not rely on periodically collecting information on user-specific attributes that may vary over time. Thus, an online learning approach that uses solely aggregate road flow data is practically viable.




\ifarxiv
\subsection{Performance Measures to Set Optimal Tolls with Incomplete Information} \label{sec:perf-measures}
\fi

We now introduce the online learning setting and the performance measures used to gauge the efficacy of a tolling policy. \ifarxiv In particular, we \fi consider a setting wherein users make trips over multiple time periods $t = 1, \ldots, T$, e.g., over multiple days, and the O-D pairs and values of time of all users at each period are drawn i.i.d. from some unknown probability distribution. That is, the O-D pair and value of time vectors $(\w^t, \v^t) = ((w_u^t)_{u \in \U}, (v_u^t)_{u \in \U})$ are drawn i.i.d. from some (unknown) distribution $\D$ with non-negative support for the value of time vector of users\footnote{For the ease of exposition, we focus on the setting when the O-D pair and value of time vectors are drawn i.i.d. from a probability distribution. However, our proposed approach can also be extended to the setting when the cost of the outside option varies over time\ifarxiv, i.e., when O-D pair, value-of-time, and outside option cost vectors $(\w^t, \v^t) = ((w_u^t)_{u \in \U}, (v_u^t)_{u \in \U}, (\lambda_u^t)_{u \in \U})$ are drawn i.i.d. from some (unknown) distribution\fi.}. Note that a special case of this involves the setting where the O-D pair and value-of-time $(w_u^t, v_u^t)$ of user $u$ at time $t$ is drawn i.i.d. from some distribution $\D_u$, and the distribution $\D = \otimes_{u \in \U} \D_u$. Under \ifarxiv the aforementioned \fi i.i.d. assumption on users' trip attributes, we focus on \emph{privacy-preserving} tolling policies, wherein the central planner makes a tolling decision using only past observations of the aggregate flows on the different roads of the traffic network. In other words, the tolling policy $\ppi = (\pi_1, \ldots, \pi_T)$ that sets a sequence of tolls $\ttau^{(1)}, \ldots, \ttau^{(T)}$ is such that $\ttau^{(t)} = \pi_t(\{\x^{t'} \}_{t'=1}^{t-1})$, where $\x^{t}$ are the aggregate edge flows corresponding to an equilibrium solution $\f^t$ under a toll $\ttau^{(t-1)}$.

We evaluate the efficacy of an algorithm $\ppi = (\pi_1, \ldots, \pi_T)$ using two metrics: (i) expected cumulative regret and (ii) expected cumulative constraint violation, where the expectation is with respect to the O-D pair and value-of-time distribution $\D$ of the users.

The regret of an algorithm $\ppi$ with a corresponding sequence of tolls $\ttau^{(1)}, \ldots, \ttau^{(T)}$ is evaluated through the expected difference between the optimal Objective~\eqref{eq:capacityObj1IP}, given complete information on the values of time and O-D pairs of all users at each time period, and the cumulative objective of the algorithm $\ppi$ over the $T$ periods. Let $\f^{t}, \f_o^t$ and $\f^{t*}, \f_o^{t*}$ denote the equilibrium flow at time $t$ given the toll $\ttau^{(t)}$ and the optimal toll $\ttau^{(t)*}$, respectively. Further, let $\P_u^t$ be the set of feasible paths corresponding to the O-D pair $w_u^t$. Then, the regret $R_T(\ppi)$ of an algorithm $\ppi$ is given by 
\ifarxiv
\begin{align*}
    R_T(\ppi) = \mathbb{E}_{\D} \left[ \sum_{t = 1}^T \left( \sum_{u \in \U} \left( v_u^t \sum_{P \in \mathcal{P}_u^t }  l_P f_{Pu}^t + \lambda_u f_{o,u}^t \right) -  \sum_{u \in \U} \left( v_u^t \sum_{P \in \mathcal{P}_u^t }  l_P f_{Pu}^{t*}  + \lambda_u f_{o,u}^{t*} \right) \right) \right],
\end{align*}
\fi
where the expectation is taken with respect to the distribution $\D$. In the remainder of this work, with slight abuse of notation, we will assume that all expectations are with respect to $\D$ and thus we drop the subscript $\D$ in the expectation. We mention here that regret measures in online learning often define regret based on the sub-optimality with respect to an optimal static action in hindsight. In contrast, we adopt a more powerful offline oracle model, wherein the oracle can vary its actions across time steps since the demand itself is random.


We evaluate the constraint violation of an algorithm $\ppi$ through the norm of the expected cumulative excess flow beyond each edge's capacity. That is, given the edge flows $\x^t$ corresponding to the equilibrium flows induced by the tolls $\ttau^{(t)}$ for each period $t$, the cumulative constraint violation vector $\v$ of an algorithm $\ppi$ is
\ifarxiv
\begin{align*}
    \v(\ppi) = \left( \sum_{t = 1}^T (\x^t - \c) \right)_+,
\end{align*}
\fi
and its expected norm is
\begin{align*}
    V_T(\ppi) = \mathbb{E} \left[ \norm{\v(\ppi)}_2 \right].
\end{align*}

We focus on jointly optimizing regret and capacity violation. Note that minimizing either one of these metrics is typically easy. For instance, the absence of tolls is likely to result in good regret guarantees since each user will solve a shortest path problem that disregards road capacity constraints. In contrast, setting large road tolls will likely reduce capacity violations but achieve a higher regret. Since there is a trade-off between regret and capacity violation, achieving good performance on both these measures under minimal assumptions is often challenging~\cite{li2020simple}. Thus, we focus on jointly optimizing performance across both metrics in this work.



\ifarxiv
\section{Online Learning Traffic Routing Algorithm} \label{sec:mainResult}
\fi

In this section, we develop an online learning algorithm that relies on only the aggregate flows on the roads of the traffic network and achieves sub-linear regret and constraint violation guarantees. To perform the regret analysis of this algorithm, we first consider the linear programming relaxation of Problem~\eqref{eq:capacityObj1IP}-\eqref{eq:capacityCon3IP} and its corresponding dual in Section~\ref{sec:lpdual}. We then present our online learning algorithm in Section~\ref{sec:gradientDescent}. 
Finally, in Section~\ref{sec:lowerbound}, we show that our algorithm achieves the optimal regret guarantee, up to constants, by establishing an upper bound on its regret and constraint violation and a lower bound on the regret of any online algorithm.

\subsection{Linear Programming Relaxation and Dual} \label{sec:lpdual}

We first present the linear programming relaxation of Problem~\eqref{eq:capacityObj1IP}-\eqref{eq:capacityCon3IP}
\begin{mini!}|s|[2]                   
    {\f, \f_o}                              
    {U^* = \sum_{u \in \U} \Big( v_u \sum_{P \in \mathcal{P}_u }  l_P f_{P,u} + \lambda_u f_{o,u} \Big), {\label{eq:capacityObj1}} } 
    {\label{eq:Eg001}}             
    {}                                
    \addConstraint{\sum_{P \in \mathcal{P}_{u}} f_{P,u} + f_{o,u}}{= 1, \quad \forall u \in \U, } {\label{eq:capacityCon1}}   
    \addConstraint{\f_o}{\geq \0, f_{P,u} \geq 0, \quad \forall P \in \P_u, u \in \U } {\label{eq:capacityCon2}}
    \addConstraint{\sum_{u \in \U} \sum_{P \in \mathcal{P}_{u} : e \in P} f_{P,u}}{  \leq c_e, \quad \forall e \in E , } {\label{eq:capacityCon3}}
\end{mini!}
where the binary allocation Constraints~\eqref{eq:capacityCon2IP} are relaxed with non-negativity Constraints~\eqref{eq:capacityCon2}. Denote $\mu_u$ as the dual variable for the allocation Constraint~\eqref{eq:capacityCon1} for each user $u$ and the toll $\tau_e$ as the dual variable of the capacity Constraint~\eqref{eq:capacityCon3} for each edge $e$. Then, the vector of road tolls $\ttau$ can be computed through the following dual of the linear Program~\eqref{eq:capacityObj1}-\eqref{eq:capacityCon3}
\begin{maxi!}|s|[2]                   
    {\tau_e, \mu_u}                              
    {\sum_{u \in \U} \mu_u - \sum_{e \in E} \tau_e c_e, \label{eq:DualOPT-Obj-Roughgarden2}}   
    {\label{eq:Eg001}}             
    {}                                
    \addConstraint{\tau_e}{\geq 0, \quad \forall e \in E, \label{eq:DualOPTcon1-Roughgarden2}}    
    \addConstraint{\mu_u}{  \leq v_u l_P + \sum_{e \in P} \tau_e, \ifarxiv \quad \fi \forall P \in \mathcal{P}_u, u \in \U, \label{eq:DualOPTcon2-Roughgarden2}}
    \addConstraint{\mu_u}{  \leq \lambda_u, \quad \forall u \in \U. \label{eq:DualOPTcon2-Roughgarden3}}
\end{maxi!}
A few comments about the dual Problem~\eqref{eq:DualOPT-Obj-Roughgarden2}-\eqref{eq:DualOPTcon2-Roughgarden3} are in order. First, the optimal tolls $\ttau^*$ satisfy a market-clearing property that $\tau_e = 0$ on a given edge if the aggregate flow on edge $e$ is strictly below capacity and $\tau_e \geq 0$ otherwise. Next, Constraint~\eqref{eq:DualOPTcon2-Roughgarden2} (and Constraint~\eqref{eq:DualOPTcon2-Roughgarden3}), together with the complimentary slackness optimality conditions, imply that the flow of user $u$ on path $P$ (or the outside option) is strictly greater than zero, i.e., $f_{P,u} > 0$ (or $f_{o,u}>0$), if the dual variable $\mu_u$ of the allocation Constraint~\eqref{eq:capacityCon1} is equal to the travel cost on that path (or the cost of the outside option). That is, Constraints~\eqref{eq:DualOPTcon2-Roughgarden2} and~\eqref{eq:DualOPTcon2-Roughgarden3} imply that the travel cost $\mu_u$ incurred by each user $u$ is the minimum across all paths and the outside option under the edge tolls $\ttau^*$. As a result, the optimal solution $\f^*, \f_o^*$ to the system optimization Problem~\eqref{eq:capacityObj1}-\eqref{eq:capacityCon3} is a (non-atomic) equilibrium under the optimal tolls $\ttau^*$ to the dual Problem~\eqref{eq:DualOPT-Obj-Roughgarden2}-\eqref{eq:DualOPTcon2-Roughgarden2}. Finally, since $\mu_u$ is the minimum travel cost across all feasible paths and the outside option for each user, the dual Problem~\eqref{eq:DualOPT-Obj-Roughgarden2}-\eqref{eq:DualOPTcon2-Roughgarden2} can be reformulated solely in terms of the tolls $\ttau$ as 
\begin{maxi!}|s|[2]                   
    {\ttau \geq \0}                              
    {\!\sum_{u \in \U} \min \! \Big\{ \! \min_{P \in \P_u} \! \Big\{ \! v_u l_P \! + \! \sum_{e \in P} \tau_e \! \Big\}, \! \lambda_u \! \Big\} \! - \! \sum_{e \in E} \! \tau_e c_e. \label{eq:DualOPT-Obj-Roughgarden3}}   
    {\label{eq:Eg001}}             
    {}                                
\end{maxi!}

\subsection{Online Learning Algorithm} \label{sec:gradientDescent}

\begin{figure}[tbh!]
    \centering
    \includegraphics[width=\linewidth]{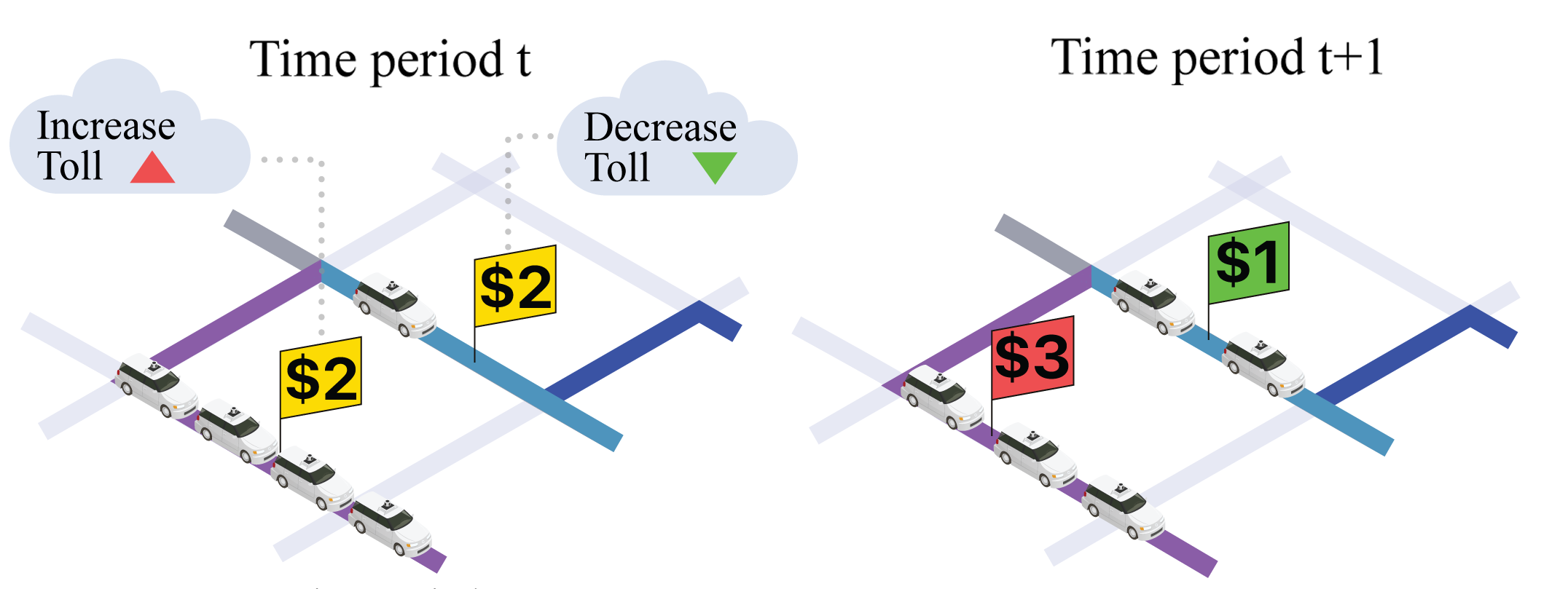}
    \vspace{-10pt}
    \caption{{\small \sf Toll update step in Algorithm~\ref{alg:GradientDescent}. Between subsequent time periods, the toll on each edge is increased (decreased) if there is more (less) flow than the capacity of that edge.}} \vspace{-15pt}
    \label{fig:tollupdate}
\end{figure}

In this section, we leverage the dual Problem~\eqref{eq:DualOPT-Obj-Roughgarden3} to derive an algorithm to dynamically set tolls for the setting when the O-D pair and values of time of users are drawn i.i.d. from some unknown distribution. In this setting, we develop an algorithm wherein the toll on each edge is adjusted at each time step based on the observed aggregate flows at the previous time step. In particular, we increase the toll on an edge if its flow is higher than its capacity and decrease the edge's toll if its flow is strictly lower than its capacity. To update the tolls, we use a step-size $\gamma$ and ensure that the tolls are non-negative at each period $t \in [T]$. We reiterate that given the toll $\ttau^{(t)}$ at each period $t$, users choose paths (or the outside option) to minimize their travel costs. This process is presented formally in Algorithm~\ref{alg:GradientDescent}, and the toll update procedure is depicted in Figure~\ref{fig:tollupdate}.


\begin{algorithm} 
\SetAlgoLined
\SetKwInOut{Input}{Input}\SetKwInOut{Output}{Output}
\Input{Time Period $T$, Road Capacities $\c$}
Set the Toll $\ttau^{(1)} \leftarrow \0 $ \\
 \For{$t = 1, \ldots, T$}{
 \textbf{Phase I: User Equilibrium for Toll $\ttau^{(t)}$} \\
 Initialize $\f^t, \f_o^t \leftarrow \0$ \\
 \tcc{Minimum cost Route}
 $Q_u^* \leftarrow \argmin_{Q \in \P_u^t \cup \{o\}} \big\{ \min_{P\in \P_u^t} \{ v_u^t l_P + \sum_{e \in P} \tau_e^{(t)} \}, \lambda_u \big\}$  \;
 \tcc{Users choose Paths}
 For each user $u \in \U$, set $f_{Q_u^*, u}^t = 1$ if $Q_u^* \in \P_u^t$, else $f_{o,u}^t = 1$ \;
 \tcc{Observed Edge Flows}
 $x_e^t \leftarrow \sum_{u \in U} \sum_{P:e \in P} f_{P, u}^t$ \;
 \vspace{3pt}
 \textbf{Phase II: Toll Update} \\
 $\ttau^{(t+1)} \leftarrow (\ttau^{(t)} - \gamma (\c - \x^t))_+$ 
 }
\caption{Efficient Routing via Privacy-Preserving Tolls} 
\label{alg:GradientDescent}
\end{algorithm} 
A few comments about Algorithm~\ref{alg:GradientDescent} are in order. First, Algorithm~\ref{alg:GradientDescent} is privacy-preserving since the toll updates do not require any information on the O-D pair, values of time of users, or their traversed paths and only rely on the observed aggregate flows on each edge of the network. Next, the computational complexity of the toll updates (Phase II of Algorithm~\ref{alg:GradientDescent}) at each period $t$ is only $O(|E|)$, which makes Algorithm~\ref{alg:GradientDescent} practically viable. Furthermore, the computational complexity of Phase I of Algorithm~\ref{alg:GradientDescent} is $O( |\U| (|E| + |V| \log(|V|)))$, since at each period $t$ each user solves a shortest path problem on the graph, which has a complexity of $O(|E| + |V| \log(|V|))$, in response to the set tolls. 
Finally, \ifarxiv we mention that \fi the toll update procedure in Algorithm~\ref{alg:GradientDescent} follows as a direct consequence of applying gradient descent to the dual Problem~\eqref{eq:DualOPT-Obj-Roughgarden3}. \ifarxiv To see this, observe that
\ifarxiv
\begin{align*}
    \partial_{\tau_e} \left( \sum_{u \in \U} \min \bigg\{ \min_{P \in \P_u } \Big\{ v_u l_P + \sum_{e \in P} \tau_e \Big\}, \lambda_u \bigg\} - \sum_{e \in E} \tau_e c_e \right) = \sum_{u \in \U} \sum_{P \in \P_u: e \in P} f_{P, u} - c_e = x_e - c_e,
\end{align*}
\else 
\begin{align*}
    &\partial_{\tau_e} \left( \sum_{u \in \U} \min \bigg\{ \min_{P \in \P_u } \Big\{ v_u l_P + \sum_{e \in P} \tau_e \Big\}, \lambda_u \bigg\} - \sum_{e \in E} \tau_e c_e \right) \\ &= \sum_{u \in \U} \sum_{P \in \P_u: e \in P} f_{P, u} - c_e = x_e - c_e,
\end{align*}
\fi
which follows since $\sum_{u \in \U} \sum_{P \in \P_u: e \in P} f_{P, u} = x_e$.
\fi


\subsection{Regret and Constraint Violation Guarantees} \label{sec:lowerbound}

We now show that Algorithm~\ref{alg:GradientDescent} achieves the optimal regret guarantee, up to constants, by establishing matching upper and lower bounds on its regret. 

We first present the main result of this work, which establishes that both the regret and constraint violation of Algorithm~\ref{alg:GradientDescent} are upper bounded by $O(\sqrt{T})$.

\begin{theorem} [Square Root Regret and Constraint Violation] \label{thm:squareRootRegret}
Suppose that the O-D pairs and values of time of users are drawn i.i.d. at each period $t \in [T]$ from some distribution $\D$ with non-negative support for users' values of time. Further, let $\x^t$ be the sequence of observed traffic flows under the equilibrium flows $\f^t, \f_o^t$ given the tolls $\ttau^{(t)}$ for each period $t$. Then, under Algorithm~\ref{alg:GradientDescent} with step size $\gamma = \frac{1}{\sqrt{T}}$, the regret 
\ifarxiv
$$R_T(\ppi) \leq \frac{|E| (|\U|+\max_{e \in E} c_e)^2}{2} \sqrt{T},$$
\fi
and constraint violation\footnote{The constant derived for the constraint violation bound is for the $L_2$ norm, and by norm equivalence this square root upper bound on the constraint violation holds for any $p$-norm, e.g., the $L_{\infty}$ norm.} 
\ifarxiv
$$V_T(\ppi) \leq |E| (\max_{u \in \U} \lambda_u + \max_{e \in E} c_e + |\U|) \sqrt{T}.$$
\fi
\end{theorem}

\begin{hproof}
To prove this claim, we first derive a generic upper bound on the regret of any algorithm using linear programming duality, and then use the toll update steps to establish an upper bound on the regret of Algorithm~\ref{alg:GradientDescent}. For the constraint violation bound, we first use the toll update steps to obtain a $O(\frac{1}{\gamma} \mathbb{E} [\norm{\ttau^{(T+1)}}_2])$ upper bound. Then, we show that the tolls remain bounded by a constant at each time step, since users will never travel on a path with a cost greater than that of their outside option. This establishes that the bound on the constraint violation is $O(\frac{1}{\gamma}) = O(\sqrt{T})$.
\qed
\end{hproof}
\ifarxiv
For a complete proof of Theorem~\ref{thm:squareRootRegret}, see Appendix~\ref{sec:proofGrDe}.
\else
\fi


Having established an upper bound on the regret of Algorithm~\ref{alg:GradientDescent} in Theorem~\ref{thm:squareRootRegret}, we now show that no algorithm can achieve a regret lower than $\Omega(\sqrt{T})$ to establish that Algorithm~\ref{alg:GradientDescent} is optimal up to constants. \ifarxiv In particular, Theorem~\ref{thm:regretlb} establishes an $\Omega(\sqrt{T})$ lower bound on the regret of any algorithm for the traffic routing problem where the O-D pair and values-of-time of users are drawn i.i.d. at each period from an unknown probability distribution $\D$. \fi

\begin{theorem} [Regret Lower Bound] \label{thm:regretlb}
There exists a distribution $\D$ such that the regret of any algorithm is $\Omega(\sqrt{T})$ for the traffic routing problem where the O-D pair and values of time of users are drawn i.i.d. at each period $t \in [T]$ from $\D$.
\end{theorem}
\ifarxiv
\begin{proof}
Consider a one edge network, where the edge $e_1 = (s_1, t_1)$ is between the origin $s_1$ and destination $t_1$, respectively, and has a capacity of one and a fixed travel time of one. Further, consider a population of two users, i.e., $|\U| = 2$, and a distribution $\D$ such that the two users have a value-of-time $v_1 = 1$ and cost of outside option $c_1 = 2$ with probability $0.5$, and parameters $v_2 = 0$ and $c_2 = 0$ with probability $0.5$. We term the first group of users as type I and the users with parameters $v_2 = 0$ and $c_2 = 0$ as type II users.'

To prove this claim, first observe that over a time horizon of $T$ days, a total of $T$ users can travel in the network, since there is a capacity of one for each day. Next, note that the optimal allocation rule during each period $t$ is to assign both users to edge $e_1$ if they are of type I, and to assign the users of type II to the outside option. To derive the $\Omega(\sqrt{T})$ regret lower bound, we suppose that $S_1$ users of type I arrive over the time horizon $T$, i.e., users of type I arrive over $\frac{S_1}{2}$ periods with two users arriving at each period. We now show that the $\Omega(\sqrt{T})$ regret for this problem arises from the fact that there may be too many arrivals of type I users into the system over the $T$ periods.

Since $T$ users of type I arrive in expectation, note that the expected cumulative regret is given by
\begin{align*}
    \text{Regret} = \mathbb{E} \left[ \max \{S_1, T\} - T \right] c_1 = 2 \mathbb{E} \left[ (S_1 - T)_+ \right],
\end{align*}
Then, from the central limit theorem, it is clear that the regret is $\Omega(\sqrt{T})$.
\end{proof}
We mention that the above proof is an application of lower bound proofs in the online linear programming literature~\cite{shamir2013complexity,hazan2014bandit} to the traffic routing context considered in this work.
\fi

\section{Numerical Experiments} \label{sec:expt}

\ifarxiv
In this section, we evaluate the performance of Algorithm~\ref{alg:GradientDescent} on a real-world transportation network. The results of our experiments not only validate the theoretical guarantees obtained in Theorem~\ref{thm:squareRootRegret} but also show that our algorithm achieves better performance on regret, constraint violation, and total travel time metrics as compared to several benchmark algorithms. In this section, we introduce three benchmark approaches to which we compare Algorithm~\ref{alg:GradientDescent} (Section~\ref{sec:benchmark}), present the data set used for the experiments, and discuss the implementation details of Algorithm~\ref{alg:GradientDescent} and the benchmarks (Section~\ref{sec:implDetails}). Then, we present the results that demonstrate the theoretical and practical efficacy of Algorithm~\ref{alg:GradientDescent} in Sections~\ref{sec:mainNumericalResults} and~\ref{sec:pracConsiderations}, respectively. Our code is publicly available at \href{https://github.com/StanfordASL/online-tolls}{https://github.com/StanfordASL/online-tolls}.
\fi

\subsection{Benchmarks} \label{sec:benchmark}

In our experiments, we compare Algorithm~\ref{alg:GradientDescent} to several benchmark approaches. In particular, the first two benchmarks assume some knowledge about the user attributes, i.e., the mean value of time of the entire population or that of each user, and set static tolls that do not vary over time. Note that setting static tolls is akin to many existing road tolling schemes. On the other hand, the third benchmark is analogous to Algorithm~\ref{alg:GradientDescent} in that it does not require any information on the values of time of users; however, in this benchmark, the tolls are updated by a fixed constant at each time step.


\ifarxiv
\paragraph{Population Value-of-Time (Population Mean VoT):} 
\fi

We assume that the central planner has access to the mean value-of-time of users in the entire population. In this setting, we compute the tolls through the optimal dual variables of the capacity constraints of Problem~\eqref{eq:capacityObj1}-\eqref{eq:capacityCon3} when the values-of-time $v_u$ of each user $u$ are set equal to the mean value of time of the entire population. Note that this benchmark does not account for the variability in users' values of time, which is reflective of many congestion pricing policies 
that neglect users' values of time in their tolling decisions.

\ifarxiv
\paragraph{Mean User Value-of-time (Group Mean VoT):} 
\fi

We assume that the central planner has more fine-grained information through access to the mean value of time of each user. In this context, we compute the tolls through the optimal dual variables of the capacity constraints of Problem~\eqref{eq:capacityObj1}-\eqref{eq:capacityCon3} when the values-of-time $v_u$ of each user $u$ are set equal to their mean value of time.

\ifarxiv
\paragraph{Reactive Toll Updates:} 
\fi

We consider a variant of Algorithm~\ref{alg:GradientDescent}, wherein the toll on all edges is updated at each period by a constant (set to $\$0.1$ in our experiments) solely based on the observed aggregate flows on the edges of the traffic network. In particular, if the flow exceeds the edge's capacity, the toll is increased by a specified constant irrespective of the magnitude of the capacity violation. On the other hand, if the flow is below the edge's capacity, the toll is decreased by a constant (or set to zero). This toll update procedure could be a natural strategy 
in large-scale traffic networks, where tolls can only be adjusted in constant increments.



\ifarxiv
\subsection{Data Set and Implementation Details} \label{sec:implDetails}

For our numerical study, we test Algorithm~\ref{alg:GradientDescent} and the benchmarks introduced in Section~\ref{sec:benchmark} on the Sioux Falls network, depicted in Figure~\ref{fig:road_network}, obtained from~\cite{tntp}. The data set contains both origin-destination travel information for all users and information on the capacity, length, and maximum speed of every road in the traffic network. To obtain the travel time of every edge, we assume that vehicles travel at the maximum speed for that edge. Furthermore, we scale the total user demand by a factor of $0.5$ to ensure feasibility of the linear Program~\eqref{eq:capacityObj1IP}-\eqref{eq:capacityCon3IP}.

Since the computational complexity of solving the linear Program~\eqref{eq:capacityObj1IP}-\eqref{eq:capacityCon3IP} scales with the number of users, we group users with the same origin-destination pairs to have the same values-of-time to improve the computational tractability of the resulting optimization problems. For the experiments, we further assume that the value-of-time for users scales proportionally to their incomes. In Sioux Falls, the range of people's incomes ranges from below $\$10,000$ to over $\$200,000$~\cite{SFIncome}, which amounts to a value-of-time range from about $\$5/$hr to $\$100/$hr, assuming 40 hours of work a week for 50 working weeks. Then, for each user group $g$, we draw their mean value-of-time, denoted by $\mu_g$, uniformly at random between the range $\$5/$hr and $\$100$/hr. We further assume that at each period, all users from a group have a value-of-time drawn from a uniform distribution over the interval $[0.8 \mu_g, 1.2 \mu_g]$. The O-D pair for every user is drawn from a distribution defined as follows: with probability 0.8, the user travels on their default O-D pair, as described in the data set, and with probability 0.2, the user chooses an O-D pair uniformly at random from the space of all possible O-D pairs. In Algorithm~\ref{alg:GradientDescent}, we set the step-size $\gamma = 5 \times 10^{-4}$. Finally, to break ties among equivalent minimum cost routes for users, we add a noise to the Population Mean VoT and User Mean VoT tolls every time step drawn from the uniform distribution in $[-5\times 10^{-4}, 5 \times 10^{-4}]$.

\ifarxiv
To efficiently implement the outside option we consider a modified network with additional edges between the corresponding O-D pairs for each user group. We set the capacity of these edges to be strictly higher than the total demand between the corresponding O-D pair. Furthermore, we set the travel time of these edges to be 1.5 times the cost, including travel times and tolls, of the shortest path under the optimal tolls. We mention that to improve computational tractability, we club the outside options for all users belonging to the same group (i.e., having the same origin-destination pair) into one edge. In this modified network, each user must traverse one path, which may be a path in the original graph or on the added edge representing the outside option for that user. Finally, note that the tolls on the added edges must be zero for all tested algorithms since the maximum possible flow on any of the added edges will be lower than the edge capacities by construction.
\else
To calibrate the cost of the outside option, we set $\lambda_u$ to be 1.5 times the cost, including travel times and tolls, of the shortest path under the optimal tolls for each user.
\fi
\fi

\ifarxiv
\begin{figure}[!ht]
    \centering
    \begin{subfigure}[b]{0.45\textwidth}
    \includegraphics[width=\linewidth]{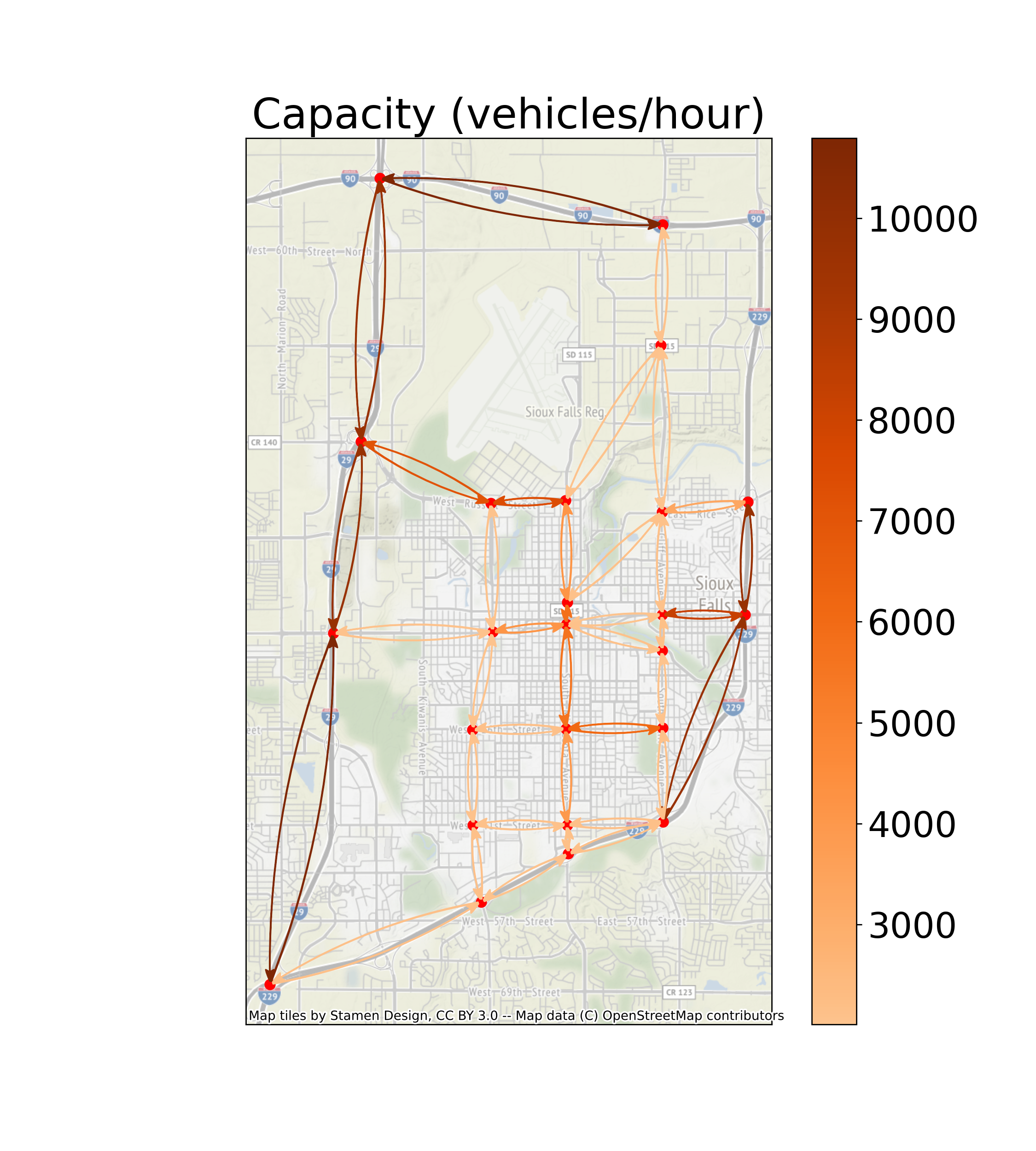}
    \end{subfigure}
    \label{fig:my_label}
    \begin{subfigure}[b]{0.45\textwidth}
    \includegraphics[width=\linewidth]{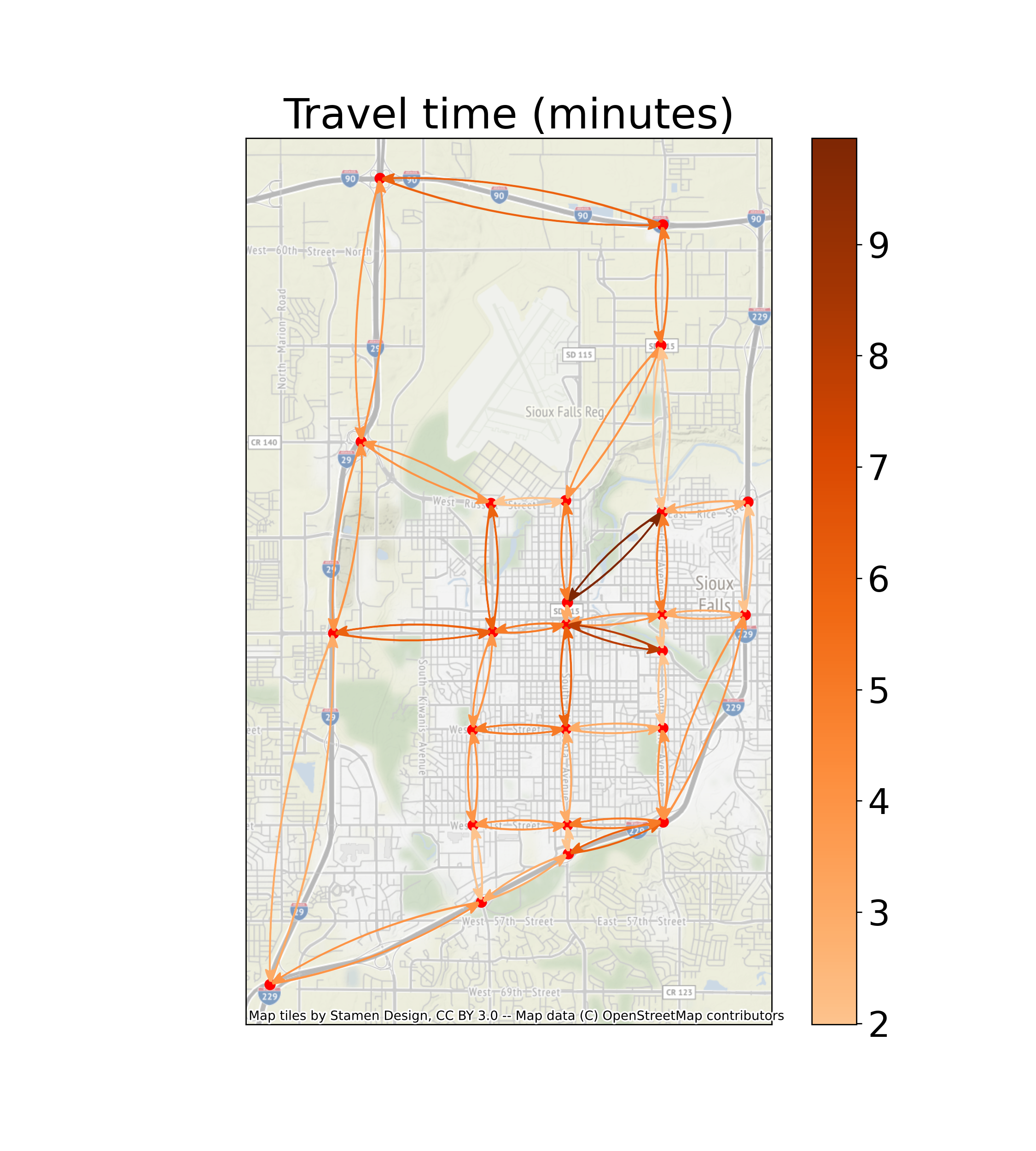}
    \end{subfigure}
    \vspace{-30pt}
    \caption{{\small \sf Depiction of the Sioux Falls road network. The capacities on each edge of the network are depicted on the left, while the travel times are depicted on the right.}}
    \label{fig:road_network}
\end{figure}
\fi
\subsection{Results} \label{sec:mainNumericalResults}

\ifarxiv
\paragraph{Assessment of Theoretical Bounds:}
\fi

\begin{figure}[tbh!]
    \centering
    \begin{subfigure}[t]{0.49\columnwidth}
        \centering
            \includegraphics[width=\linewidth]{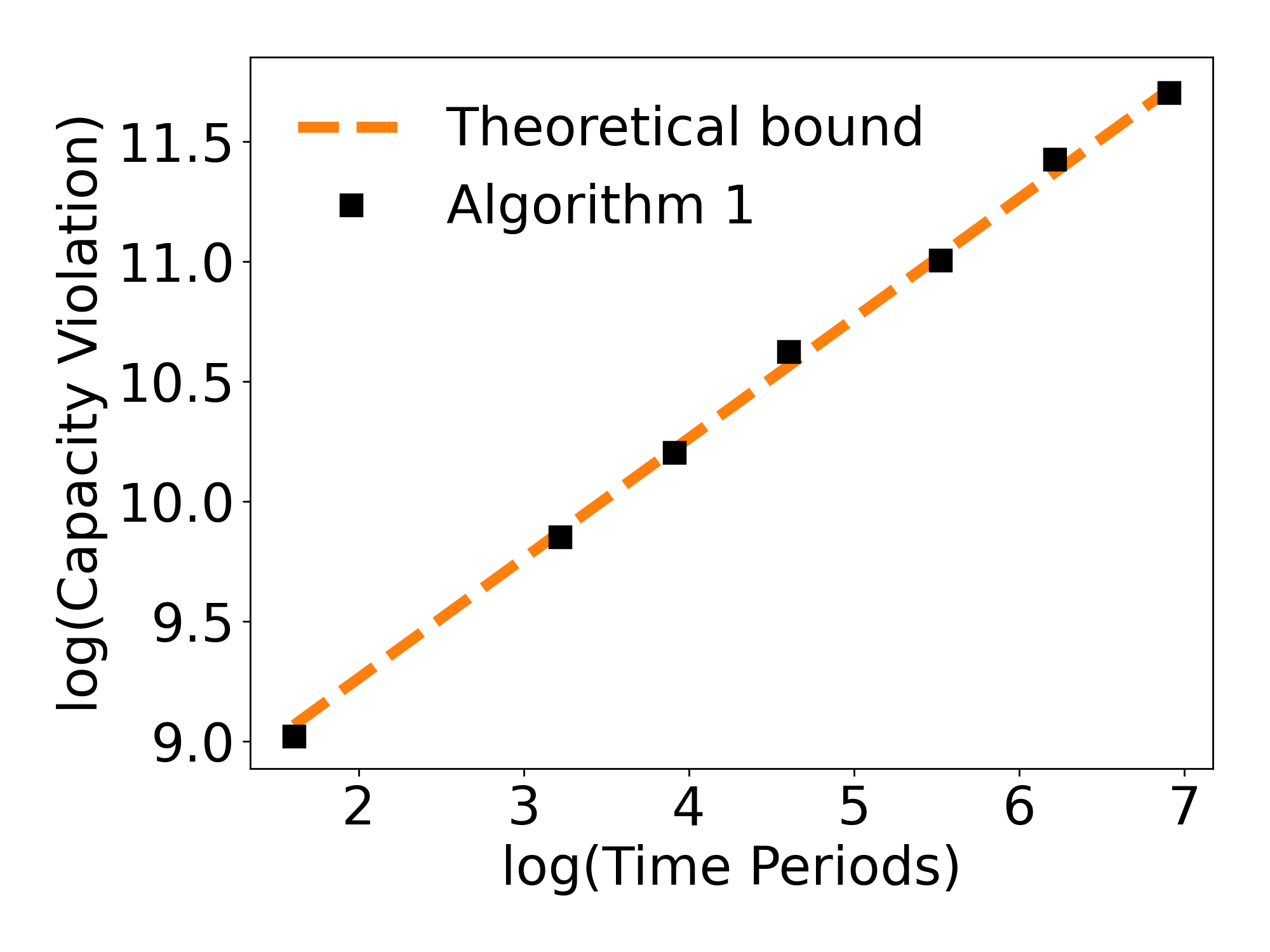}
    \end{subfigure} 
    \begin{subfigure}[t]{0.49\columnwidth}
        \centering
            \includegraphics[width=\linewidth]{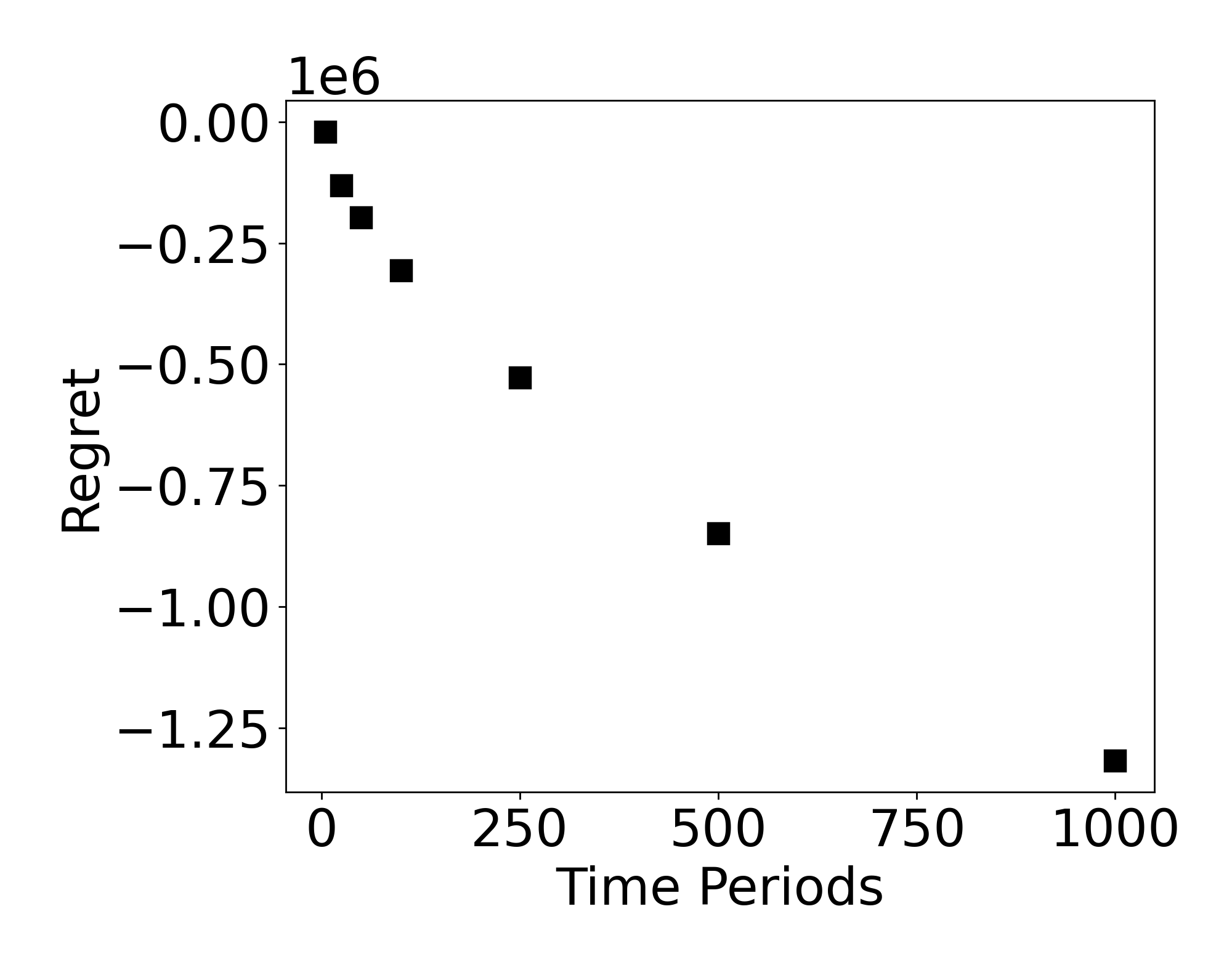}
    \end{subfigure}
    \caption{{\small \sf Validation of the theoretical upper bounds on regret and constraint violation obtained in Theorem~\ref{thm:regretlb} on the Sioux Falls data set. The infinity norm of the capacity violation is plotted on a log-log plot, and it can be seen that the empirical performance is very close, with a root mean square error of 0.037, to the theoretical $O(\sqrt{T})$ bound, which is represented by a line with a slope of $0.5$. The regret is negative for this data set due to capacity violations and thus trivially satisfies the $O(\sqrt{T})$ bound.}} 
    \label{fig:theoryvallidate}
\end{figure}

Figure~\ref{fig:theoryvallidate} depicts the regret (right) and a log-log plot of the capacity violation (left), wherein we consider the setting when both the O-D pairs and values of time of users are drawn i.i.d. from a fixed distribution. As expected from our theoretical results, for the capacity violation, the black dots representing the empirically observed capacity violations of Algorithm~\ref{alg:GradientDescent} in Figure~\ref{fig:theoryvallidate} all lie very close to the theoretical $O(\sqrt{T})$ bound represented by a line with a slope of $0.5$ on a logarithmic scale. Furthermore, the regret also satisfies the $O(\sqrt{T})$ bound since it is negative for this data set due to capacity violations.


\ifarxiv
\begin{figure}[!ht]
      \centering
      \includegraphics[width=0.9\linewidth]{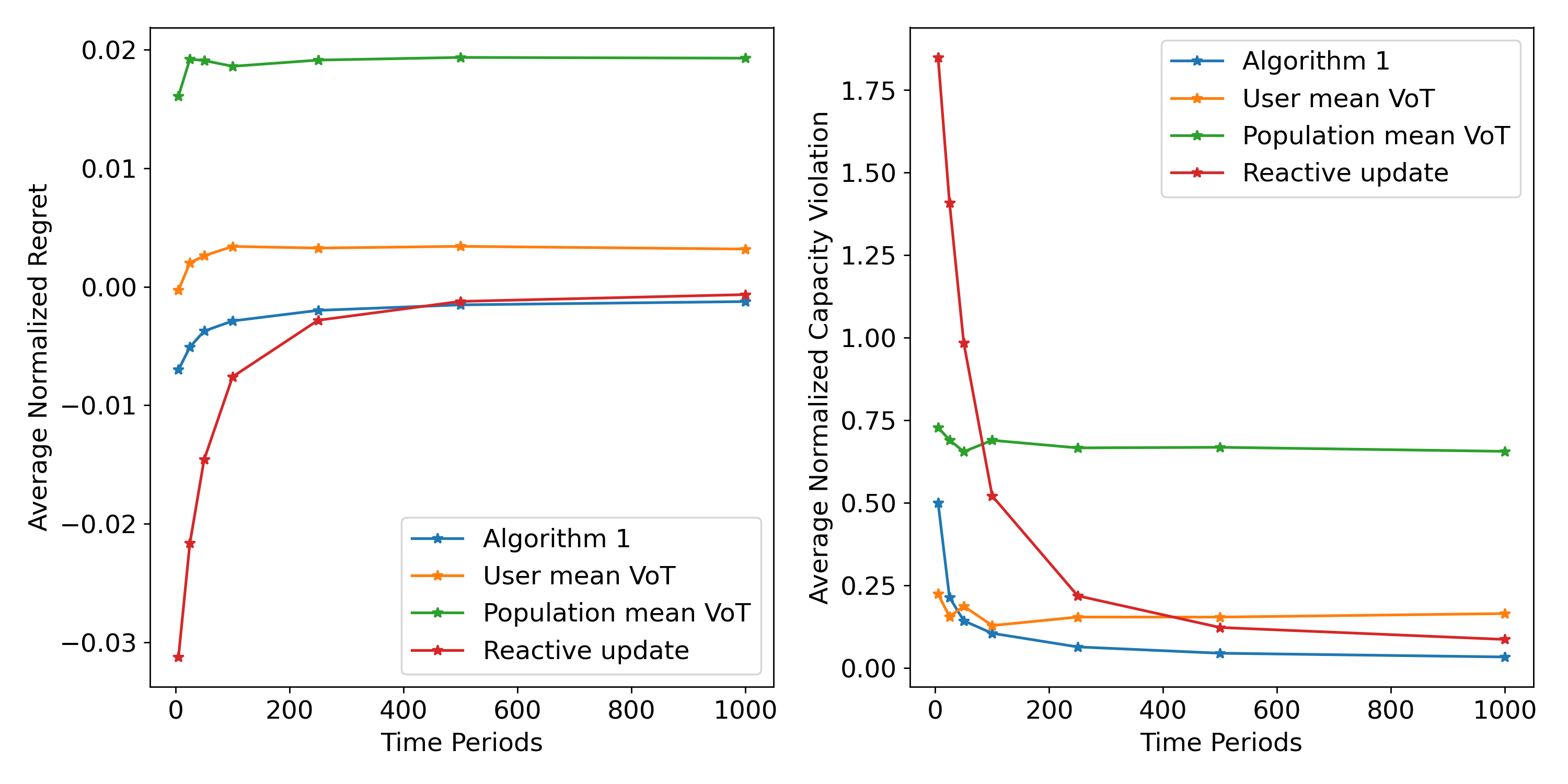}
      \caption{{\small \sf Comparison of the normalized regret and capacity violation of Algorithm~\ref{alg:GradientDescent} and the benchmark algorithms presented in Section~\ref{sec:benchmark}. The normalized regret represents the ratio between the regret and the optimal offline total system cost over the $T$ time periods, while the normalized capacity violation is the ratio between the $L_{\infty}$ norm of the capacity violation capacity violation and the cumulative road capacity over the $T$ periods.}}
      \label{fig:convergenceplot}
   \end{figure}
   
\fi

\ifarxiv
For the remainder of the experiments, we assume that the O-D pair for each user does not change with time to improve the computational tractability of the simulations. Observe that this setting is a special case where the i.i.d. distribution for the O-D pairs collapses to a point mass.
\fi

\ifarxiv
\paragraph{Regret and Capacity Violation Comparisons:}
\fi

\ifarxiv
We now compare the normalized regret and capacity violation of Algorithm~\ref{alg:GradientDescent} to that of the benchmark algorithms in Figure~\ref{fig:convergenceplot}. Here, the normalized regret is the ratio between the regret and the optimal offline total system cost over the $T$ time periods, and the normalized capacity violation is the ratio between the $L_{\infty}$ norm of the capacity violation and the cumulative capacity over the $T$ periods. We mention that we consider the $L_{\infty}$ norm of the capacity violation here since this involves the computation of the maximum capacity violation across all roads in the traffic network. In Figure~\ref{fig:convergenceplot} (left, center) we observe that Algorithm~\ref{alg:GradientDescent} (i) outperforms all the benchmark algorithms on both metrics for large values of $T$, (ii) obtains significantly better performance in terms of regret as compared to the two static toll benchmarks for all values of $T$, and (iii) obtains a superior performance in terms of constraint violation as compared to the Reactive Toll Updates benchmark for all values of $T$.
\fi

Between the two static tolling benchmarks, we observe from Figure~\ref{fig:convergenceplot} that the User Mean VoT benchmark performs better than the Population Mean VoT benchmark on both regret and constraint violation metrics. This result follows since the User Mean VoT benchmark has access to fine-grained information on the values of time of each user while the Population Mean VoT benchmark only has access to the mean value of time of the entire population. The performance of these two benchmarks, thus, points to the importance of considering the variability in users' values of time in designing congestion pricing schemes.

Compared to the static tolling benchmarks that assume knowledge of the mean values of time of each user (or the population), both dynamic tolling policies do not have access to any information on users' trip attributes. Despite this, we observe from Figure~\ref{fig:convergenceplot} that both the dynamic tolling policies achieve better regret guarantees as compared to the static tolling benchmarks. The primary reason for the low regret guarantee for the Reactive Toll Update benchmark is that it has a higher level of capacity violation that enables users to take lower-cost routes and thus achieve a lower level of regret. For larger values of the number of periods $T$, Figure~\ref{fig:convergenceplot} indicates that Algorithm~\ref{alg:GradientDescent} eventually outperforms the static toll benchmarks on both regret and capacity violation. This observation suggests that setting fixed tolls, even using the mean values of time of each user, might be fraught with error since users' values of time may vary over time, even though the distribution from which values of time are drawn is stationary.



Between the two dynamic tolling policies, we observe that the Reactive Toll Update benchmark achieves a lower regret for small values of $T$ because of higher violations of the edge capacities. However, for larger values of $T$, Algorithm~\ref{alg:GradientDescent} outperforms this benchmark on both the regret and capacity violation metrics. This result follows since Algorithm~\ref{alg:GradientDescent} updates tolls based on the exact discrepancy between the capacity and the flow on the edges of the network. On the other hand, the Reactive Toll Update benchmark updates the tolls on each edge by a pre-specified increment depending solely on whether the capacity is violated. As a result, Algorithm~\ref{alg:GradientDescent}, which can make infinitesimally small toll updates, achieves a vanishing normalized regret and capacity violation with large $T$. However, we do note that the Reactive Toll Update procedure does come close to the performance of Algorithm~\ref{alg:GradientDescent} since it achieves only a slightly higher regret and only an 8\% capacity violation for large values of $T$. Such a performance indicates that the Reactive Toll Update, in addition to Algorithm~\ref{alg:GradientDescent}, can also be practically deployed in real-world transportation networks.

\ifarxiv
\subsection{Other Practical Considerations} \label{sec:pracConsiderations}

In this section, we demonstrate the practical efficacy of Algorithm~\ref{alg:GradientDescent} by comparing its total travel time, which may also be an important practical consideration for a central planner, to that of the benchmark algorithms and investigating the properties of the tolls set in the traffic network.

\paragraph{Total Travel Time:}


Figure~\ref{fig:convergenceplot} (right) depicts the ratio of the average total travel time of each of the algorithms to the minimum achievable total travel time in the network, which corresponds to a solution satisfying the capacity constraints of all roads. In particular, both the dynamic tolling policies achieve lower total travel times than the static tolling benchmarks. Furthermore, while incurring small levels of capacity violation, Algorithm~\ref{alg:GradientDescent} achieves close to the minimum possible total travel time in the network. Thus, even though Theorem~\ref{thm:squareRootRegret} only provides guarantees for Algorithm~\ref{alg:GradientDescent} on regret and constraint violation metrics, it achieves good practical performance on even the total travel time metric, which may be of direct importance to transportation planners.

\ifarxiv
\begin{figure}[!ht]
      \centering
      \includegraphics[width=0.5\linewidth]{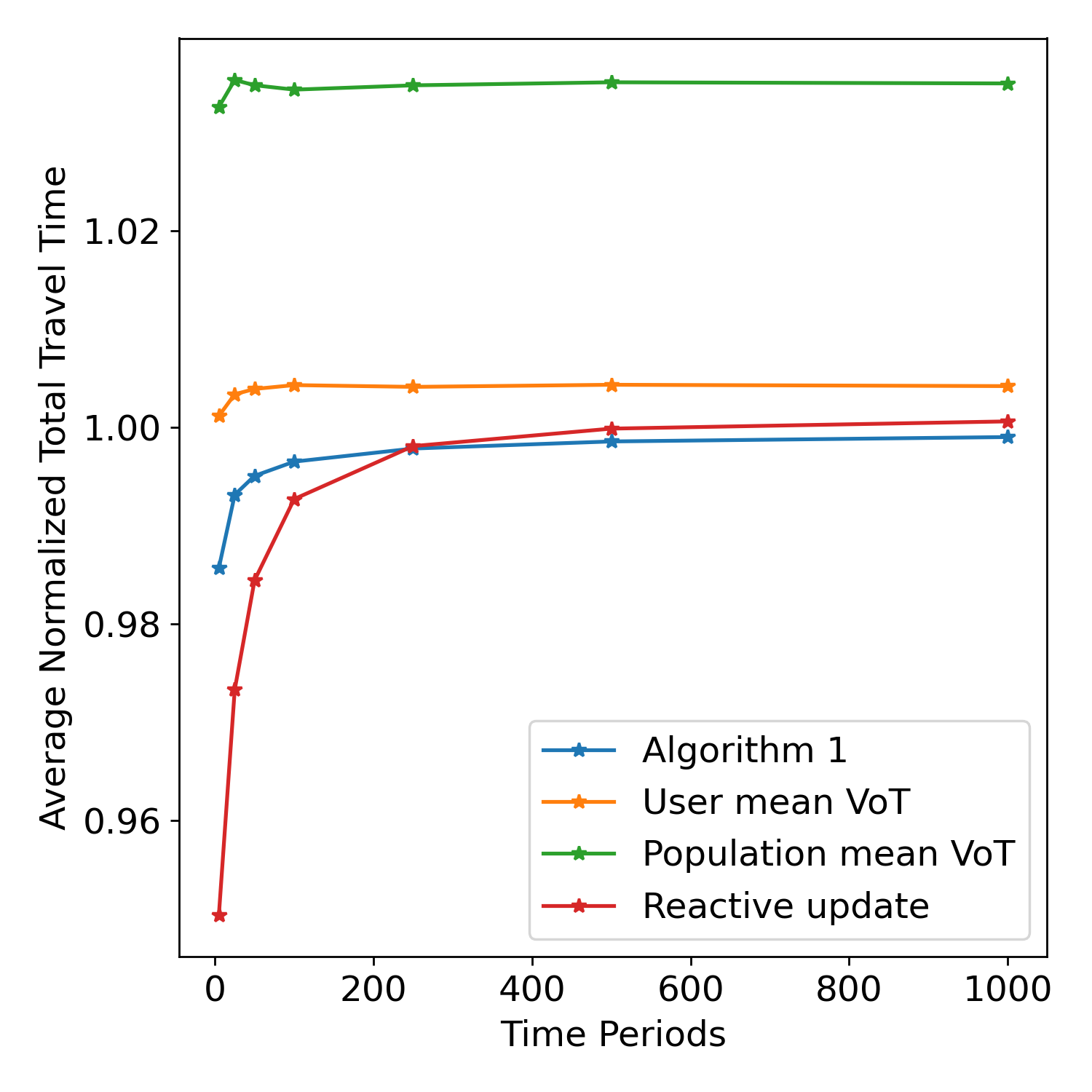}
      \caption{{\small \sf Comparison of the average total travel time of Algorithm~\ref{alg:GradientDescent} and the benchmark algorithms presented in Section~\ref{sec:benchmark}. The y-axis represents the ratio between the average total travel time corresponding to each algorithm over $T$ periods and the optimal total travel time. Here, the optimal total travel time represents the minimum total travel time of a traffic routing solution that satisfies the road capacity constraints.}}
      \label{fig:tttPlot}
   \end{figure}
\fi


\paragraph{Properties of Computed Tolls:}

\ifarxiv
\begin{figure}
    \centering
    \begin{subfigure}[b]{0.35\textwidth}
    \includegraphics[width=\linewidth]{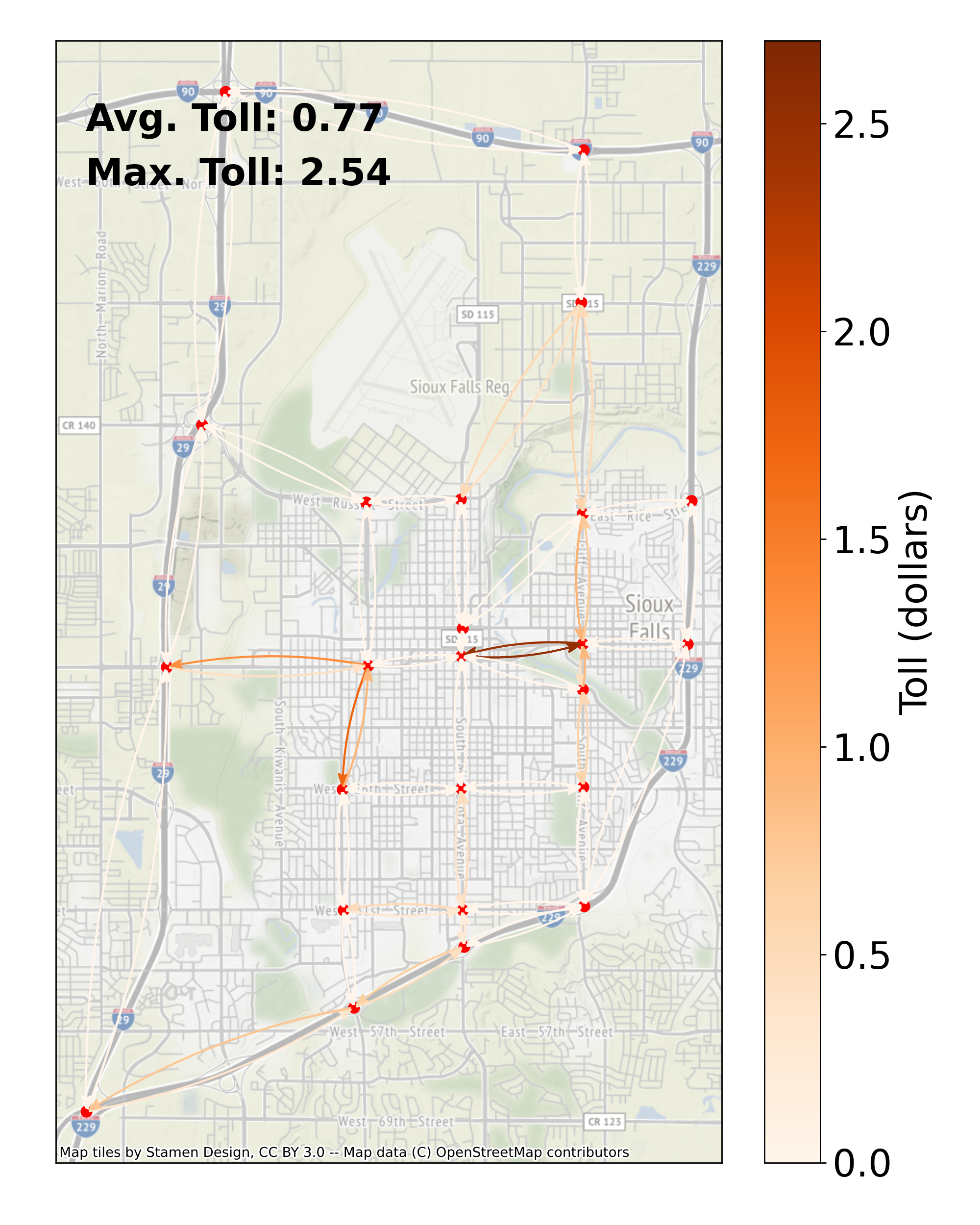}
    \subcaption{Algorithm 1}
    \end{subfigure}
    \begin{subfigure}[b]{0.35\textwidth}
    \includegraphics[width=\linewidth]{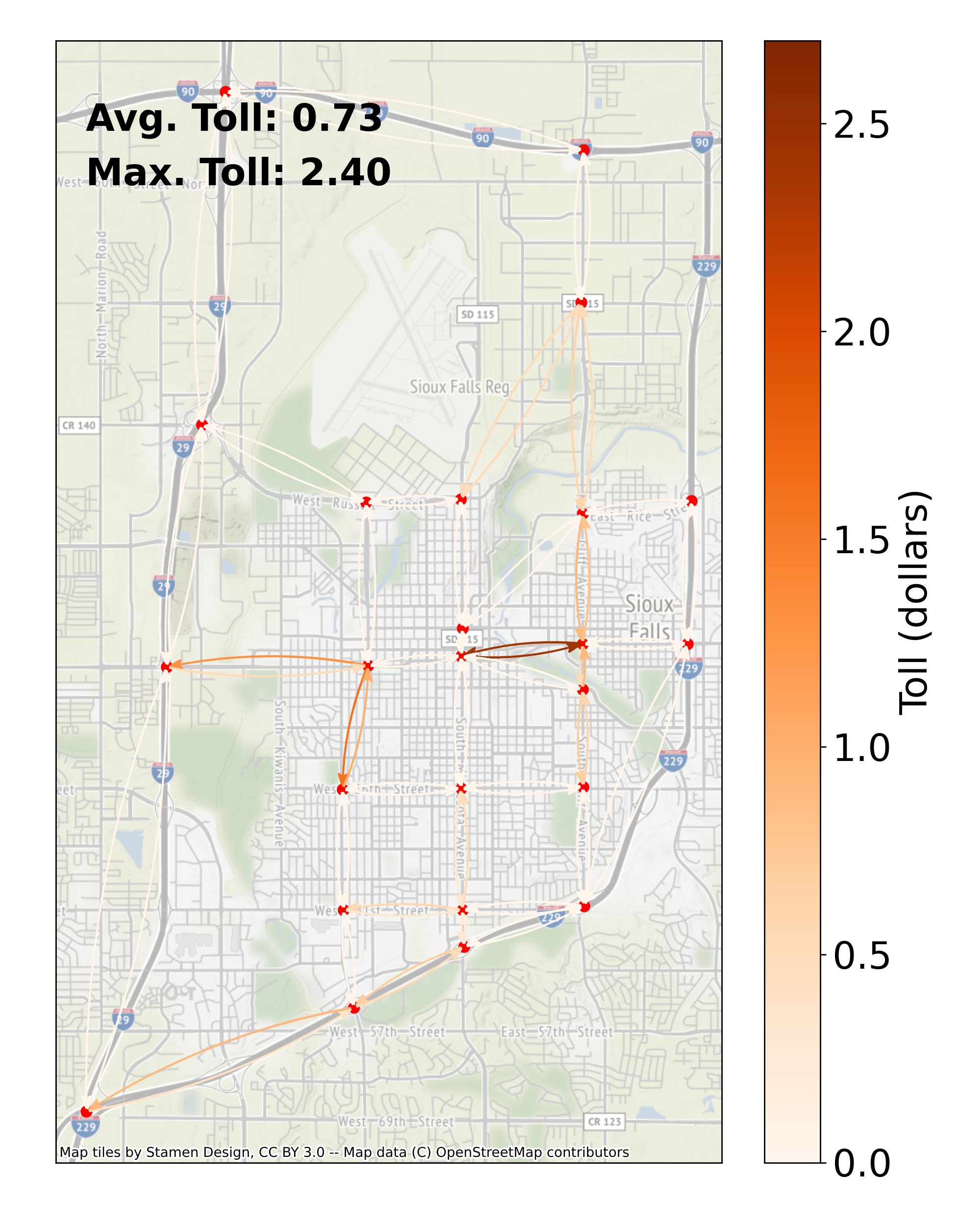}
    \subcaption{Reactive update}
    \end{subfigure}
    \begin{subfigure}[b]{0.35\textwidth}
    \includegraphics[width=\linewidth]{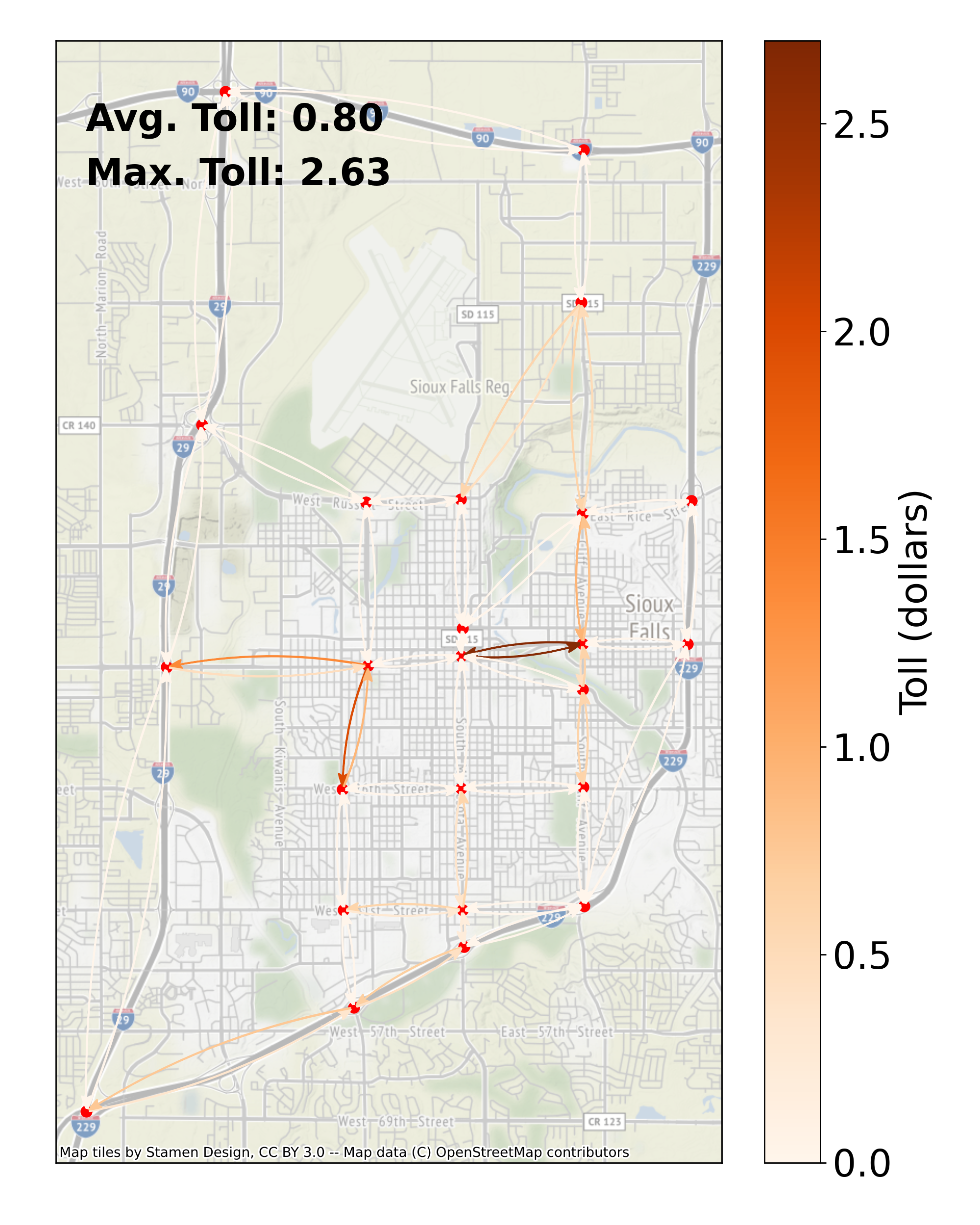}
    \subcaption{User mean VoT}
    \end{subfigure}
    \begin{subfigure}[b]{0.35\textwidth}
    \includegraphics[width=\linewidth]{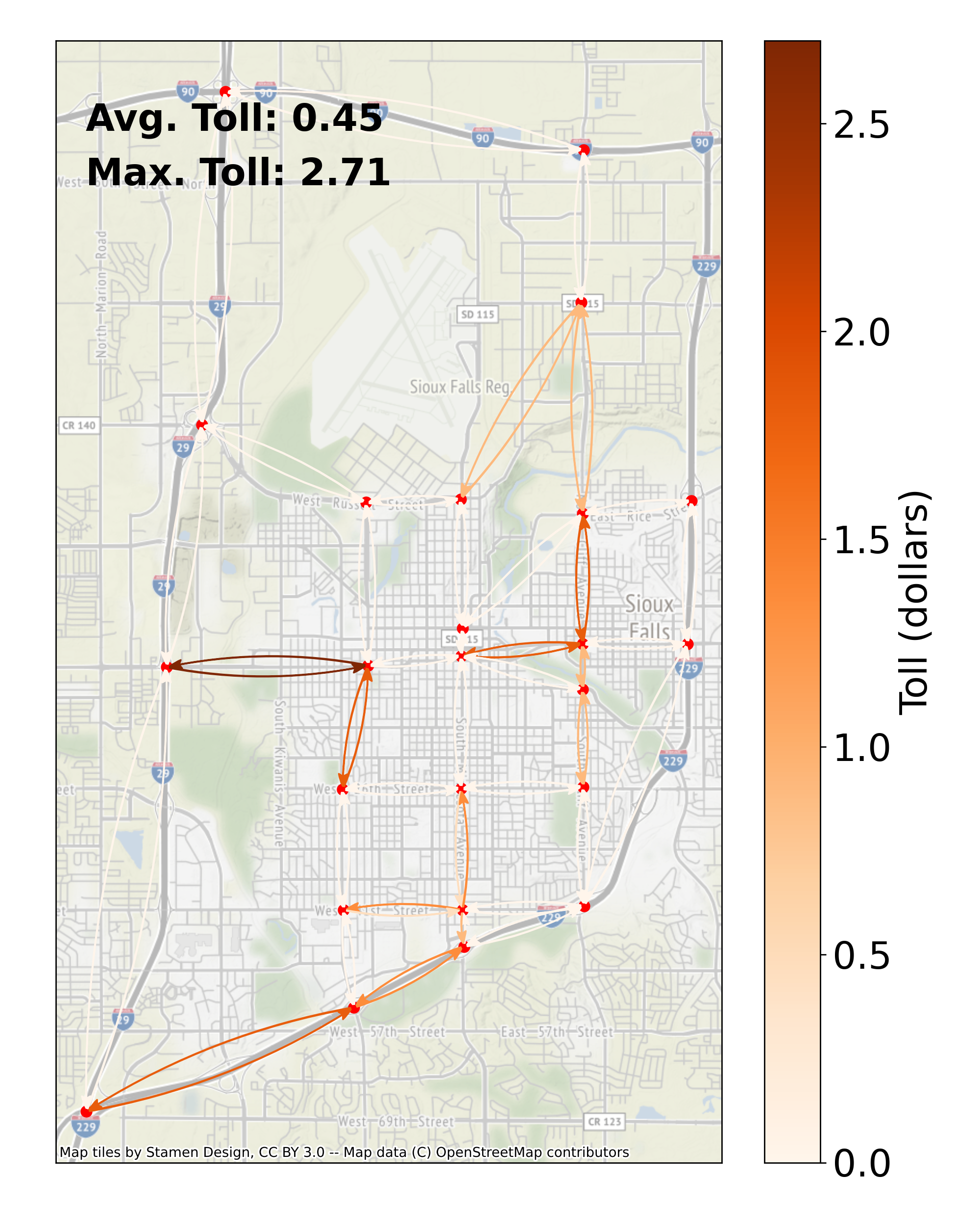}
    \subcaption{Population mean VoT}
    \end{subfigure}
    \caption{{\small \sf Tolls corresponding to Algorithm~\ref{alg:GradientDescent} and the three benchmark approaches in the Sioux Falls traffic network at $T = 1000$.}}
    \label{fig:tollRangePlot}
\end{figure}
\else

\fi

We now present some properties of the tolls set using the Algorithm~\ref{alg:GradientDescent} and the benchmarks. To this end, Figure~\ref{fig:tollRangePlot} depicts the tolls set by Algorithm~\ref{alg:GradientDescent} and the three benchmarks at $T = 1000$. From this figure, we observe that the tolls are typically placed either on roads in the dense urban areas on the center-right of the Sioux Falls network or on roads on the left and bottom of the network that have smaller road capacities. We further note that the set tolls were about $\$0.75$ on average for the two dynamic algorithms and the User Mean VoT benchmark, with the maximum toll for the algorithms ranging between $\$2.4$-$\$2.7$, which is in alignment with the order of magnitude of typical tolls in real-world transportation networks.

In Figure~\ref{fig:toll_hist}, we present a histogram representing the edge tolls corresponding to the Algorithm~\ref{alg:GradientDescent} at $T = 1000$. This histogram indicates that most road tolls are zero, which corresponds to traditional congestion pricing schemes that operate on only certain roads or regions of the traffic network. Furthermore, we observe that most non-zero tolls are in the range of $\$0.25$-$\$1.00$, while only about five percent of the edges have road tolls that exceed $\$1$.

Finally, in Figure~\ref{fig:convergence}, we depict the evolution of the cumulative tolls, i.e., the sum of the tolls on all edges, over the $T=1000$ time horizon. This figure illustrates that both the dynamic tolling algorithms, i.e., Algorithm~\ref{alg:GradientDescent} and the Reactive Toll Update benchmark, approach a set of road tolls in a small number of periods, after which the tolls oscillate to achieve a good balance between constraint violation and regret. Note that the Reactive Toll Update benchmark stabilizes to a small range of toll values earlier with its constant toll updates as compared to Algorithm~\ref{alg:GradientDescent} since the step size of the updates of Algorithm~\ref{alg:GradientDescent} is of the order $O(\frac{1}{\sqrt{1000}})$ for $T = 1000$. However, due to the constant updates in the tolls at each time step, the Reactive Toll Update benchmark also has a greater degree of variability in its tolls after arriving at a stable value for the cumulative tolls.

\ifarxiv
\begin{figure}[!htb]
    \centering
    \begin{minipage}{.45\textwidth}
        \centering
        \includegraphics[width=0.9\linewidth]{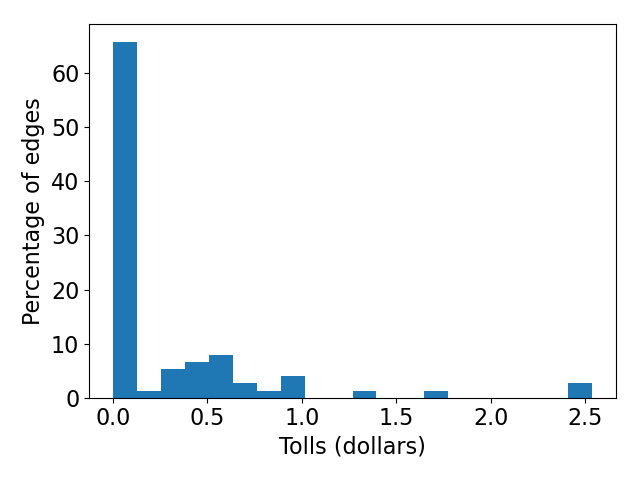}
        \caption{{\small \sf Histogram of edge tolls in the Sioux Falls traffic network for Algorithm~\ref{alg:GradientDescent} at T=1000.}}
    \label{fig:toll_hist}
    \end{minipage} \hspace{5pt}
    \begin{minipage}{0.45\textwidth}
        \centering
        \includegraphics[width=0.9\linewidth]{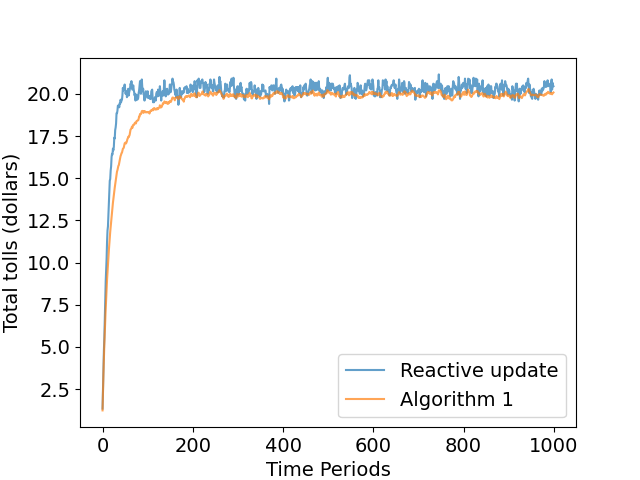}
        \caption{{ \small \sf Time evolution of cumulative tolls on all edges of the Sioux Falls traffic network for Algorithm~\ref{alg:GradientDescent} and the Reactive Toll Update benchmark for $T = 1000$.}}
        \label{fig:convergence}
    \end{minipage}
\end{figure}

\else
\begin{figure}
    \centering
    \begin{subfigure}[t]{0.49\columnwidth}
        \centering
            \includegraphics[width=\linewidth]{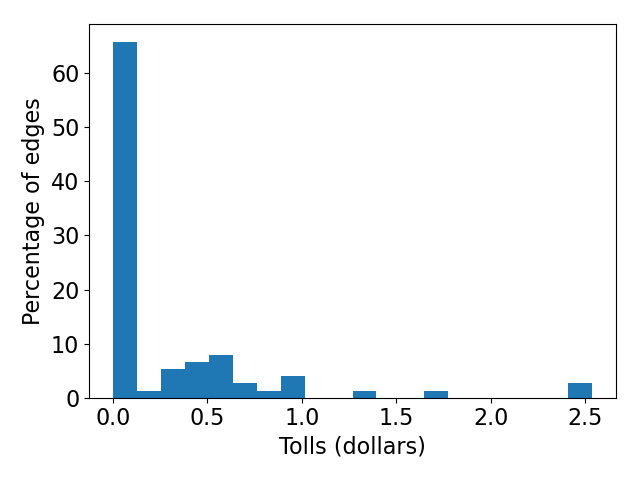}
    \end{subfigure}
    \begin{subfigure}[t]{0.49\columnwidth}
        \centering
        \includegraphics[width=\linewidth]{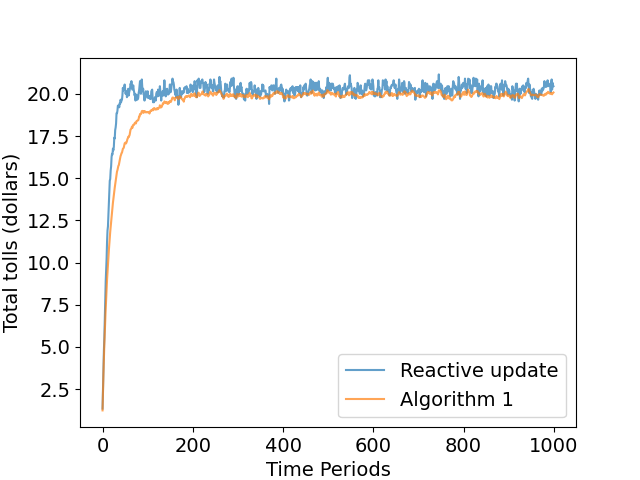}
    \end{subfigure}
    \caption{Histogram of edge tolls in the Sioux Falls traffic network for Algorithm~\ref{alg:GradientDescent} at T=1000. Time evolution of cumulative tolls on all edges of the Sioux Falls traffic network for Algorithm~\ref{alg:GradientDescent} and the Reactive Toll Update benchmark for $T = 1000$.}
    \label{fig:convergence}
\end{figure}
\fi

\fi

\section{Conclusion and Future Work} \label{sec:conclusion}

In this work, we proposed an online learning approach to set tolls in a traffic network to induce heterogeneous users with different values of time toward a system-optimum traffic pattern. \ifarxiv 
In particular, we developed an online learning algorithm that adjusts road tolls based solely on the observed aggregate flows on the edges of the traffic network and involves a simple toll update rule that is easy to compute in practice. We further showed that this algorithm achieves sub-linear regret and constraint violation guarantees (in the number of periods $T$ over which the tolls are updated) that are optimal up to constants. Finally, we evaluated the performance of our approach on a real-world transportation network, which highlighted both the theoretical and practical efficacy of our online learning algorithm to set road tolls. \else \fi

There are various directions for future research. First, it would be interesting to investigate whether the regret and capacity violation guarantees extend to settings when the users' trip attributes are not drawn i.i.d., e.g., when their trip attributes are drawn according to a random permutation model~\cite{li2020simple}. Next, it would be worthwhile to generalize the results obtained in this work to the context of congestion games, where the travel time on each edge is a function of its flow. Finally, it would also be valuable to explore the extension of the ideas and algorithm developed in this paper to objectives beyond system efficiency, e.g., achieving fairness or maximizing revenue.

\ifarxiv
\section*{Acknowledgements}

This research was supported by the National Science Foundation under the CPS program, the Center for Automotive Research at Stanford University, and Docomo Inc. We thank Emma Brunskill, Michael Ostrovsky, Dorsa Sadigh, Edward Schmerling, Kiril Solovey, and Matthew Tsao for helpful discussions, and Matteo Zallio for his help in crafting Figure~\ref{fig:tollupdate}.
\else 

\fi

\bibliographystyle{unsrt}
\bibliography{main}

\ifarxiv
\appendix

\section{Proof of Theorem~\ref{thm:squareRootRegret}} \label{sec:proofGrDe}

In this section, we present the proof of Theorem~\ref{thm:squareRootRegret}. To this end, we first present a generic bound on the regret of any algorithm and then use this bound to derive an upper bound on the regret of Algorithm~\ref{alg:GradientDescent} in terms of the step size $\gamma$. We then derive an upper bound on the constraint violation of Algorithm~\ref{alg:GradientDescent} in terms of the step size $\gamma$ as well. Finally, choosing $\gamma = O(\frac{1}{\sqrt{T}})$, we obtain that both these regret and constraint violation bounds are $O(\sqrt{T})$.

\subsection{Generic Bound on Regret} \label{sec:genericBound}

We first present an upper bound on the expected regret of any algorithm that we will use to derive upper bounds on the regret of Algorithm~\ref{alg:GradientDescent}.

\begin{lemma} [Generic Regret Bound] \label{lem:generic-regretBound}
Consider an algorithm $\ppi$ that sets a sequence of tolls $\ttau^{(t)}$ and let $\f^t, \f_o^t$ be the resulting equilibrium flows for each time period $t \in [T]$. Then, denoting $\x^t$ as the sequence of observed traffic flows corresponding to the equilibrium flows $\f^t$ for each time period $t \in [T]$, the regret
\begin{align*}
    R_T(\ppi) \leq \mathbb{E} \left[ \sum_{t = 1}^T \ttau^{(t)} \cdot (\c - \x^t) \right].
\end{align*}
\end{lemma}

\begin{proof}
To prove this claim, we first present a lower bound on the expected optimal objective $\mathbb{E} \left[ U_t^* \right]$ at each time $t \in [T]$. Then, we present an upper bound on the expected regret accrued at each time $t \in [T]$ to obtain the desired regret upper bound.

For each $t \in [T]$, let $\f^{t*},\f_o^{t*}$ be the optimal solution of Problem~\eqref{eq:capacityObj1}-\eqref{eq:capacityCon3} and $\ttau^{(t)*}$ be the optimal tolls. Then, observe for each $t \in [T]$ that
\begin{align*}
    \mathbb{E} \left[ U_t^* \right] &= \mathbb{E} \left[ \sum_{u \in \U} \left( v_u^t \sum_{P \in \mathcal{P}_u^t}  l_P f_{Pu}^{t*} + \lambda_u f_{o,u}^{t*} \right) \right], \\
    &\stackrel{(a)}{=} \mathbb{E} \left[ \sum_{u \in \U} \min \bigg\{ \min_{P\in \P_u^t} \Big\{ v_u^t l_P + \sum_{e \in P} \tau_e^{(t)*} \Big\}, \lambda_u \bigg\} - \sum_{e \in E} \tau_e^{(t)*} c_e \right], \\
    &\stackrel{(b)}{\geq} \mathbb{E} \left[ \sum_{u \in \U} \min \bigg\{ \min_{P\in \P_u^t} \Big\{ v_u^t l_P + \sum_{e \in P} \tau_e^{(t)} \Big\}, \lambda_u \bigg\} - \sum_{e \in E} \tau_e^{(t)} c_e \right],
\end{align*}
where (a) follows by strong duality and (b) follows by the optimality of $\ttau^{(t)*}$ for the dual Problem~\eqref{eq:DualOPT-Obj-Roughgarden3}.

Next, we let $\f^t,\f_o^t$ denote the vectors that encode the equilibrium flows under the tolls $\ttau^{(t)}$, i.e., the flow $f_{Pu}^t$ ($f_{o,u}^t$) denotes whether user $u$ was routed on $P$ (the outside option) at time period $t$. Then, letting the objective $U_t = \sum_{u \in \U} v_u^t \left( \sum_{P \in \mathcal{P}_u^t}  l_P f_{Pu}^t + \lambda_u f_{o,u}^t \right)$ be the total system cost incurred under the toll $\ttau^{(t)}$, we have from the above lower bound on the expected optimal objective $\mathbb{E} \left[ U_t^* \right]$ that
\begin{align*}
    \mathbb{E} \left[ U_t - U_t^* \right] &\leq \mathbb{E} \left[ \sum_{u \in \U} v_u^t \left( \sum_{P \in \mathcal{P}_u^t}  l_P f_{Pu}^t + \lambda_u f_{o,u}^t \right) - \sum_{u \in \U} \min \bigg\{ \min_{P\in \P_u^t} \Big\{ v_u^t l_P + \sum_{e \in P} \tau_e^{(t)} \Big\}, \lambda_u \bigg\} + \sum_{e \in E} \tau_e^{(t)} c_e \right], \\
    &\stackrel{(a)}{=} \mathbb{E} \left[ \sum_{u \in \U} v_u^t \left( \sum_{P \in \mathcal{P}_u^t} \! l_P f_{Pu}^t \! + \! \lambda_u f_{o,u}^t \right) \! + \! \sum_{e \in E} \tau_e^{t} x_e^t \right] \\ &+\mathbb{E} \left[ \! - \! \sum_{u \in \U} \min \bigg\{ \! \min_{P\in \P_u^t} \! \Big\{ v_u^t l_P \! + \! \sum_{e \in P} \tau_e^{(t)} \Big\}, \! \lambda_u \bigg\} \! + \! \sum_{e \in E} \! \tau_e^{(t)} (c_e \! - \! x_e^t) \right], \\
    &\stackrel{(b)}{=} \mathbb{E} \left[ \sum_{e \in E} \tau_e^{(t)} (c_e - x_e^t) \right],
\end{align*}
where (a) follows by adding and subtracting the term $\sum_{e \in E} \tau_e^{(t)} x_e^t$, and (b) follows from the observation that the total travel cost $\sum_{u \in \U} v_u^t \left( \sum_{P \in \mathcal{P}_u^t}  l_P f_{Pu}^t + \lambda_u f_{o,u}^t \right) + \sum_{e \in E} \tau_e^{(t)} x_e^t$ of users at equilibrium under a toll $\ttau^{(t)}$ is exactly $\sum_{u \in \U} \min \{ \min_{P\in \P_u^t} \{ v_u^t l_P + \sum_{e \in P} \tau_e^{(t)} \}, \lambda_u \}$ since each user minimizes their corresponding travel cost.

Summing the above inequality over all time periods $t \in [T]$, we get that
\begin{align*}
    R_T(\ppi) = \mathbb{E} \left[ \sum_{t = 1}^T (U_t - U_t^*) \right] \leq \mathbb{E} \left[ \sum_{t = 1}^T \ttau^{(t)} \cdot (\c - \x^t) \right],
\end{align*}
which proves our claim.
\end{proof}

\subsection{Upper Bound on Regret of Algorithm~\ref{alg:GradientDescent}}



We now use the generic upper bound on the expected regret to show that the regret of Algorithm~\ref{alg:GradientDescent} is upper bounded by $O(\gamma T)$.

\begin{lemma} [Upper Bound on Regret] \label{lem:regretUpperBound}
Let $\x^t$ be the sequence of observed traffic flows under the equilibrium flows $\f^t, \f_o^t$ given the tolls $\ttau^{(t)}$ for each $t \in [T]$ under Algorithm~\ref{alg:GradientDescent}. Then for Algorithm~\ref{alg:GradientDescent}, the regret
\begin{align*}
    R_T(\ppi) \leq \gamma T \frac{|E| (\max_{e \in E} c_e + |\U|)^2}{2}.
\end{align*}
\end{lemma}

\begin{proof}
To prove this claim, we present an upper bound on $\mathbb{E} \left[ \sum_{t = 1}^T \ttau^{(t)} \cdot (\c - \x^t) \right]$ and then use Lemma~\ref{lem:generic-regretBound} to obtain a bound on the expected regret of Algorithm~\ref{alg:GradientDescent}.

First observe from the toll update process that
\begin{align*}
    \norm{\ttau^{(t+1)}}^2 &\leq \norm{\ttau^{(t)} - \gamma (\c - \x^t)}^2, \\
    &= \norm{\ttau^{(t)}}^2 + \gamma^2 \norm{\c - \x^t}^2 - 2 \gamma \ttau^{(t)} \cdot (\c - \x^t).
\end{align*}
Rearranging this equation, summing over $t \in [T]$ and taking expectations, we get that
\begin{align*}
    \mathbb{E} \left[ \sum_{t = 1}^T \ttau^{(t)} \cdot (\c - \x^t) \right] &\leq \mathbb{E} \left[ \sum_{t = 1}^T \frac{1}{2 \gamma} (\norm{\ttau^{(t)}}^2 - \norm{\ttau^{(t+1)}}^2) + \sum_{t = 1}^T \frac{\gamma}{2} \norm{\c - \x^t}^2 \right], \\
    &\leq \mathbb{E} \left[ \frac{1}{2 \gamma} \norm{\ttau^{(1)}}^2 + \sum_{t = 1}^T \frac{\gamma}{2} \norm{\c - \x^t}^2 \right], \\
    &\stackrel{(a)}{=} \mathbb{E} \left[ \sum_{t = 1}^T \frac{\gamma}{2} \norm{\c - \x^t}^2 \right], \\
    &\stackrel{(b)}{\leq} \gamma T \frac{|E| (\max_{e \in E} c_e + |\U|)^2}{2},
\end{align*}
where (a) follows since the initial toll $\ttau^{(1)} = \0$, and (b) follows as $|c_e - x_e^t| \leq \max_{e \in E} c_e + |\U|$. Finally, applying Lemma~\ref{lem:generic-regretBound}, we have proved the desired bound on the regret of Algorithm~\ref{alg:GradientDescent}. 
\end{proof}

\subsection{Upper Bound on Constraint Violation of Algorithm~\ref{alg:GradientDescent}}

We now establish an upper bound on the constraint violation of Algorithm~\ref{alg:GradientDescent} in terms of the step size $\gamma$. To this end, we first show that the expected constraint violation of Algorithm~\ref{alg:GradientDescent} is upper bounded by $O \left(\frac{1}{\gamma} \mathbb{E} \left[ \norm{\ttau^{(T+1)}}_2 \right] \right)$ in Lemma~\ref{lem:constraintViolation}. Then, in Lemma~\ref{lem:tollsBounded}, we show that the tolls remain bounded at each time step to establish that the constraint violation of Algorithm~\ref{alg:GradientDescent} is bounded by $O(\frac{1}{\gamma})$.


\paragraph{Upper Bound on Expected Constraint Violation in terms of Step Size $\gamma$:}
We first show that the expected constraint violation of Algorithm~\ref{alg:GradientDescent} is upper bounded by $O \left(\frac{1}{\gamma} \mathbb{E} \left[ \norm{\ttau^{(T+1)}}_2 \right] \right)$.

\begin{lemma} [Constraint Violation Bound] \label{lem:constraintViolation}
Let $\x^t$ be the sequence of observed traffic flows under the equilibrium flows $\f^t, \f_o^t$ given the tolls $\ttau^{(t)}$ for each $t \in [T]$ under Algorithm~\ref{alg:GradientDescent}. Then for Algorithm~\ref{alg:GradientDescent}, the constraint violation
\begin{align*}
    V_T(\ppi) \leq \frac{1}{\gamma} \mathbb{E} \left[ \norm{\ttau^{(T+1)}}_2 \right].
\end{align*}
\end{lemma}

\begin{proof}
By the toll update process, we know that
\begin{align*}
    \ttau^{(t+1)} \geq \ttau^{(t)} - \gamma (\c - \x^t).
\end{align*}
Rearranging the above equation and summing over $t \in [T]$, we get that
\begin{align*}
    \sum_{t = 1}^T (\x^t - \c) \leq \frac{1}{\gamma} \sum_{t = 1}^T (\ttau^{(t+1)} - \ttau^{(t)}) \leq \frac{1}{\gamma} \ttau^{(T+1)}.
\end{align*}
From this, we obtain that the expected constraint violation
\begin{align*}
    V_T(\ppi) = \mathbb{E} \left[ \norm{ \left(\sum_{t=1}^{T}\left(\x^{t}-\c \right)\right)_{+}}_2 \right] \leq \frac{1}{\gamma} \mathbb{E} \left[ \norm{\ttau^{(T+1)}}_2 \right],
\end{align*}
which proves our claim.
\end{proof}

\paragraph{Boundedness of Tolls:}

Since the constraint violation is bounded by $\frac{1}{\gamma} \mathbb{E} \left[ \norm{\ttau^{(T+1)}}_2 \right]$, we seek an upper bound on the toll at time $T+1$ to obtain an upper bound for the constraint violation. In particular, we show that the tolls are bounded by a constant and thus the toll on any edge does not increase with the number of time periods $T$.

\begin{lemma} [Boundedness of Tolls] \label{lem:tollsBounded}
Under Algorithm~\ref{alg:GradientDescent}, the toll on any given edge at each time step $t$ is upper bounded by $\max_{u \in \U} \{ \lambda_u \} + \max_{e \in E} \{ c_e \} + |\U|$ for any step-size $\gamma \leq 1$.
\end{lemma}

\begin{proof}
To prove this claim, first note that the toll $\tau_e^{(1)} = 0$ for all edges and suppose that $\tau_e^{(t)} > \max_{u \in \U} \{ \lambda_u \}$. Then, it is clear that $x_e^t = 0$ and the toll on this edge must reduce in the next time step as users can use the outside option and incur a cost of $\lambda_u$ instead. Next, if $\tau_e^{(t)} \leq \max_{u \in \U} \{ \lambda_u \}$, then it must hold that $\tau_e^{(t+1)} \leq \tau_e^{(t)} + \gamma (c_e + |U|) \leq \max_{u \in \U} \{ \lambda_u \} + \max_{e \in E} \{ c_e \} + |\U|$, which proves our claim. 
\end{proof}

\subsection{Square Root Bound on Regret and Constraint Violation}


From Lemma~\ref{lem:regretUpperBound} we observed that the regret is upper bounded by $ \gamma T \frac{|E| (\max_{e \in E} c_e + |\U|)^2}{2}, i.e., O(\gamma T),$ and from Lemmas~\ref{lem:constraintViolation} and~\ref{lem:tollsBounded} we have that the expected constraint violation is upper bounded by $|E| (\max_{u \in \U} \lambda_u + \max_{e \in E} c_e + |\U|) \frac{1}{\gamma}, i.e., O(\frac{1}{\gamma}),$ since $\norm{\ttau^{(T+1)}}_2$ is bounded by a constant. Setting $\gamma T = \frac{1}{\gamma}$, we have that both the upper bounds on regret and constraint violation are minimized when $\gamma = \frac{1}{\sqrt{T}}$ as in the statement of Theorem~\ref{thm:squareRootRegret}. Finally, taking $\gamma = \frac{1}{\sqrt{T}}$, it is clear that the expected regret $R_T(\ppi) \leq \frac{|E| (\max_{e \in E} c_e + |\U|)^2}{2} \sqrt{T}$ and that the expected constraint violation $V_T(\ppi) \leq |E| (\max_{u \in \U} \lambda_u + \max_{e \in E} c_e + |\U|) \sqrt{T}$, which proves Theorem~\ref{thm:squareRootRegret}.

\section{VCG and Optimal Tolls} \label{sec:vcgOptimal}

In this section, we study whether it is possible to truthfully elicit user's preferences to determine the tolls to set in the traffic network that would induce the system optimal assignment as an equilibrium flow. To this end, we first establish a connection between the Vickrey-Clarkes-Groves (VCG) mechanism~\cite{vickrey1961counterspeculation} and tolls computed through the solution of the linear Program~\eqref{eq:DualOPT-Obj-Roughgarden2}-\eqref{eq:DualOPTcon2-Roughgarden2} to show that truthful elicitation of user's values-of-time is indeed possible in the setting of a parallel traffic network, wherein multiple parallel edges connect a single origin and destination. This result holds since the optimal market clearing toll on each edge is exactly equal to the VCG payment for users on that edge. However, for more general traffic networks we show that VCG payments of users may not correspond to market clearing tolls and thus the truthful elicitation of preferences is, in general, not possible. To simplify our analysis, we assume that there is no outside option for users, and thus all users must be routed in the traffic network.


\subsection{VCG in Parallel Road Networks}

We begin by considering the setting of parallel road networks and showing that if edge tolls are computed through the VCG mechanism, then the resulting equilibrium traffic assignment is exactly the system optimal assignment.

\begin{theorem} [VCG Tolls in Parallel Networks] \label{thm:vcg-tolls}
In a parallel network with integral capacities, the VCG mechanism induces the system optimal assignment, i.e., the solution to Problem~\eqref{eq:capacityObj1IP}-\eqref{eq:capacityCon3IP}, as an equilibrium.
\end{theorem}

\begin{proof}
To prove this claim, we first note that the VCG mechanism, by definition, is efficient and thus the VCG allocation exactly corresponds to the system optimal assignment. We now show that the VCG payments of users correspond to appropriate road tolls that induce the system optimal assignment as an equilibrium. To this end, we show that the VCG payments of all users assigned to the same edge are equal, and that if the tolls are set equal to the VCG payments of users on the corresponding edges then these tolls will induce the system optimal assignment as an equilibrium. We first introduce some notation, then present the VCG payments of users, and finally show that these VCG payments can be translated into tolls that are a solution to the Dual Problem~\eqref{eq:DualOPT-Obj-Roughgarden2}-\eqref{eq:DualOPTcon2-Roughgarden2}. Note here that, in the setting of a parallel network with integral capacities, the solution of the integer Program~\eqref{eq:capacityObj1IP}-\eqref{eq:capacityCon3IP} and the relaxed linear Program~\eqref{eq:capacityObj1}-\eqref{eq:capacityCon3} coincide.

For notational convenience, we let the edges of an $|E|$ edge parallel network be ordered by the travel times such that $l_1 \leq \ldots \leq l_{|E|}$. Furthermore, we let the users be ordered by their values-of-time such that $v_1 \geq \ldots \geq v_{|\U|}$. Then, it is clear that the system optimum solution is such that the first $c_1$ users are allocated to edge one, the next $c_2$ users are allocated to edge two, and so on. We further suppose, without loss of generality, that the user with the lowest value-of-time is routed on edge $|E|$.

We now present the VCG payments of users, which we denote by a vector $\p = \{ p_u \}_{u \in \U}$. In particular, the VCG payment captures the externality imposed by a given user on others, and thus must be the same for all users routed on a given edge, as can be observed through the following VCG payment for a user $u$ routed on edge $e$ under the system optimal assignment:
\begin{align*}
    p_u = \sum_{e':e \leq e'< M} v_{\Bar{e}'}(l_{e^{'} + 1} - l_{e^{'}}),
\end{align*}
where $\Bar{e}' = \sum_{e^{''}: e^{''} \leq e'} c_{e^{''}} + 1$ is the index of the first user using edge $e'+1$ under the system optimal assignment. Observe also that the VCG payment for all users routed on the last edge, i.e., edge $|E|$, is zero since the externality imposed by these users on others is zero.

Finally, we show that VCG payments can be translated into tolls that are a feasible solution to the Dual Problem~\eqref{eq:DualOPT-Obj-Roughgarden2}-\eqref{eq:DualOPTcon2-Roughgarden2}. In particular, consider the tolls $\tau_e = p_u$ for any user $u$ routed on edge $e$ under the system optimal assignment. We show that these tolls are a feasible solution to the Problem~\eqref{eq:DualOPT-Obj-Roughgarden2}-\eqref{eq:DualOPTcon2-Roughgarden2}. To this end, first observe that the tolls $\tau_e \geq 0$ for all edges since the payments by all users is non-negative. Next, observe that $\tau_e = \tau_{e+1} + v_{\Bar{e}} (l_{e+1} - l_e)$ for all edges $e \in E$. Then, for any user $u$ assigned to edge $e$ under the system optimal assignment it holds that
\begin{align*}
    v_u l_e + \tau_e = v_u l_e + \tau_{e+1} + v_{\Bar{e}} (l_{e+1} - l_e) \leq v_u l_{e+1} + \tau_{e+1},
\end{align*}
since $v_u \geq v_{\Bar{e}}$. The above relationship implies that a user assigned to edge $e$ under the system optimal assignment with the aforementioned VCG-based tolls incurs a lower travel cost when using edge $e$ as compared to any edge $e+1, \ldots, |E|$. We further observe for $e \geq 2$ that
\begin{align*}
    v_u l_e + \tau_e = v_u l_e + \tau_{e-1} - v_{\Bar{e}-1} (l_{e} - l_{e-1}) \leq v_u l_{e-1} + \tau_{e-1},
\end{align*}
since $v_u \leq v_{\Bar{e}-1}$. The above relationship implies that a user assigned to edge $e$ under the system optimal assignment with the aforementioned VCG-based tolls incurs a lower travel cost when using edge $e \geq 2$ as compared to any edge $1, \ldots, e-1$. The above analysis implies that the VCG based tolls induce the system optimal assignment as an equilibrium, which establishes our claim.
\end{proof}

Theorem~\ref{thm:vcg-tolls} implies that the VCG mechanism can be used to set road tolls such that the resulting equilibrium assignment is the system optimum assignment. That is, users will truthfully reveal their values-of-time to the central planner that can then set market clearing tolls that are a solution to Problem~\eqref{eq:DualOPT-Obj-Roughgarden2}-\eqref{eq:DualOPTcon2-Roughgarden2}.


\subsection{VCG in General Road Networks} \label{apdx:vcg}

In the parallel network setting, each path corresponds to a distinct edge in the graph and thus the VCG prices for each user corresponded directly to the market clearing tolls. In more general networks, however, paths may consist of more than one edge and two distinct paths may have multiple common edges. In this section, we show that the path based VCG prices cannot, in general, be decomposed into a set of edge tolls that induce the system optimal assignment as an equilibrium. To this end, we first derive a formula for the VCG payment of any user and then use this payment rule to show that the corresponding edge tolls will not, in general, result in the system optimal assignment forming an equilibrium outcome. 

We first characterize the form of the VCG payment rule for any user $u$ in general road networks through the following lemma.

\begin{lemma} \label{lem:vcgPayment}
Let $\v = \{v_u \}_{u \in \U}$ be the vector of values-of-time of users, $P^*_u(\v)$ be the path taken by user $u$ under the system optimal assignment and $P^*_u(\v_{-\Bar{u}})$ be the path taken by user $u$ under the system optimal assignment without user $\Bar{u}$. Then, in general road networks the VCG payment for a user $u \in \U$ is given by $$p_u = \sum_{\Bar{u} \in \U \backslash u} v_{\Bar{u}} (l_{P^*_{\Bar{u}}(v_{-u})} - l_{P^*_{\Bar{u}}(v)} )$$.
\end{lemma}

\begin{proof}
Let the system optimal objective be denoted as $U^*(\v) = \sum_{\Bar{u} \in \U} v_{\Bar{u}} l_{P_{\Bar{u}}^*(\v)}$ and the system optimal objective without user $u$ be $U^*(\v_{-u}) = \sum_{ \Bar{u} \in \U \backslash u} v_{\Bar{u}} l_{P_{\Bar{u}}^*(\v_{-u})}$. Then, by the VCG payment formula, we have that the payment $p_u$ for user $u$ is given by
\begin{align*}
    p_u = -[ U^*(\v_{-u}) - (U^*(\v) - v_u l_{P^*_u(\v)})].
\end{align*}
Substituting the expressions for $U^*(\v_{-u})$ and $U^*(\v)$, the above VCG payment formula reduces to
\begin{align*}
    p_u = \sum_{\Bar{u} \in \U \backslash u} v_{\Bar{u}} (l_{P^*_{\Bar{u}}(\v_{-u})} - l_{P^*_{\Bar{u}}(\v)} ),
\end{align*}
which proves our result.
\end{proof}

We now show that edge tolls corresponding to the VCG payments do not, in general, induce the system optimal assignment as an equilibrium. To this end, we consider the network in Figure~\ref{fig:countereg} where two users need to be routed between a single O-D pair $(v_1, v_6)$ where edges $e_1$ and $e_3$ have a capacity of one while the other edges have a capacity larger than two. For this problem instance, we show that tolls corresponding to VCG payments do not induce the system optimal assignment as an equilibrium. 

\begin{figure}[!ht]
      \centering
      \includegraphics[width=0.8\linewidth]{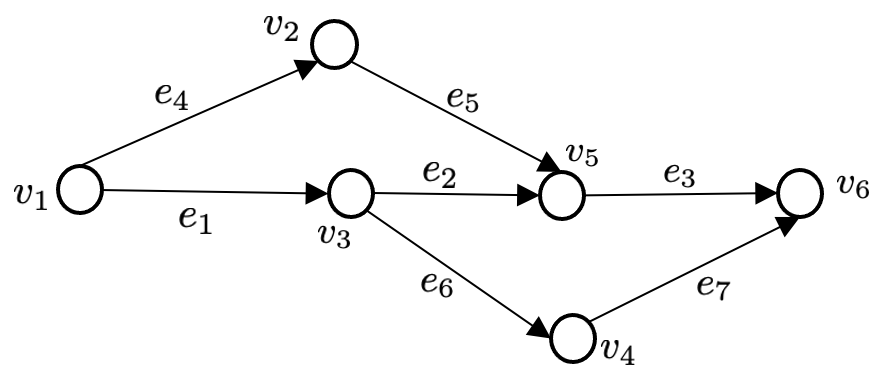}
      \caption{A single O-D pair problem instance for which the tolls corresponding to the VCG payments do not induce the system optimal assignment as an equilibrium. In this example, two users, one with a higher value-of-time than the other, need to be routed between O-D pair $(v_1, v_6)$ where edges $e_1$ and $e_3$ have a capacity of one while the other edges have a capacity larger than two.}
      \label{fig:countereg}
   \end{figure}

\begin{lemma}
The tolls corresponding to the VCG payments for the network instance in Figure~\ref{fig:countereg} does not induce the system optimal assignment, i.e., the solution to Problem~\eqref{eq:capacityObj1IP}-\eqref{eq:capacityCon3IP}, as an equilibrium.
\end{lemma}

\begin{proof}
For the problem instance in Figure~\ref{fig:countereg} there are three paths on which users can be routed, which include $P_1 = e_1-e_2-e_3$, $P_2 = e_4-e_5-e_3$ and $P_3 = e_1-e_6-e_7$. We further assume that the travel times on the paths are ordered such that $l_{P_1} < l_{P_2} < l_{P_3}$. Given the edge capacity constraints, note that the optimal solution in this context is to route the user with the higher value-of-time $v_2$ to path $P_2$ and the user with the lower value-of-time $v_3$ to path $P_3$.

Then, we have by Lemma~\ref{lem:vcgPayment} that the VCG payments for the two users are
\begin{align*}
    &p_1 = v_3(l_{P_3} - l_{P_1}), \text{ and} \\
    &p_2 = v_2(l_{P_2} - l_{P_1}).
\end{align*}
Next, observe that the only edges on which tolls need to be set are $e_1$ and $e_3$. Since user three is routed on path $P_3$ under the system optimal assignment, we can normalize the toll on edge $e_1$ to $\tau_1 = v_3(l_{P_3} - l_{P_1})$, thereby matching the VCG payment for this user. Now, we show that the toll of $\tau_3 = v_2(l_{P_2} - l_{P_1})$ on edge $e_3$ will not result in an equilibrium. In particular note that
\begin{align*}
    v_3(l_{P_3} - l_{P_2}) + v_2(l_{P_2} - l_{P_1}) - v_3(l_{P_3} - l_{P_1}) = (v_2 - v_3) (l_{P_2} - l_{P_1}) > 0
\end{align*}
Rearranging this equation implies that
\begin{align*}
    v_3 l_{P_2} + v_3(l_{P_3} - l_{P_1}) < v_3 l_{P_3} + v_2(l_{P_2} - l_{P_1}),
\end{align*}
which suggests that the under these tolls the cost on path two is lower for user two, i.e., the user with the lower value of time $v_3$, as compared to that on path three. As a result, these VCG based tolls do not induce the system optimal assignment as an equilibrium. 
\end{proof}
This result suggests that for general road networks VCG payments will, in general, not induce the system optimum solution as an equilibrium.
\fi

\end{document}